\documentclass[journal]{IEEEtai}

\usepackage[colorlinks,urlcolor=blue,linkcolor=blue,citecolor=blue]{hyperref}
\usepackage{color,array}
\usepackage{graphicx}
\usepackage{subfigure}
\usepackage{booktabs} %
\usepackage{amsmath}
\usepackage{amssymb}
\usepackage{relsize}
\usepackage{caption}
\usepackage{bbm}
\usepackage{hyperref}
\usepackage{changes}
\def \R{\mathbb{R}}
\def \M{\mathbf{M}}

\def \X{\mathbf{X}}
\def \e{\mathbf{e}}

\def \w{\mathbf{w}}
\def \xp{\mathbf{x}_{\mathbf{p}}}

\def \wi{\mathbf{w}_i}
\def \N{\mathbb{N}}
\def \sig{\sigma}

\def \ypi{y_{\mathbf{p},i}}
\def \ypk#1{y_{\mathbf{p},#1}}
\def \yp{\mathbf{y_p}}
\def \xp{\mathbf{x_p}}
\def \p{\mathbf{p}}
\def \loss{\mathcal{L}}
\def \averagep{\mathlarger{\mathbb{E}_{\mathbf{p}}}}
\def \logtwo{\log_2}
\def \qpi{q_{\mathbf{p},i}}
\def \qp{\mathbf{q_p}}

\def \K{\mathcal{K}}

\def \am{\alpha_{\mu}}
\def \nm{n_{\mu}}
\def \I{\mathcal{I}}
\newcommand{\floor}[1]{\lfloor #1 \rfloor}

\def \averagei{\mathlarger{\mathbb{E}_{i\in \I}}}
\def \L{\mathbf{L}}
\def \k{\mathbf{k}}
\def \X{\mathbf{X}}

\def \cstar{c^{\star}}

\def \Score#1{\mathcal{S}^#1}
\def \Ktr{\mathcal{K}_{train}}
\def \Kvl{\mathcal{K}_{val}}

\definecolor{darkgreen}{rgb}{0.0, 0.5, 0.0}
\definecolor{darkred}{rgb}{0.5, 0.0, 0.0}
\definecolor{softgreen}{rgb}{0.0, 0.8, 0.0}
\definecolor{purple}{RGB}{135,65,187}
\definecolor{darkgray}{rgb}{0.3, 0.3, 0.3}

\begin{document}

\title{Unsupervised Interpretable Basis Extraction for Concept -- Based Visual Explanations} 

\author{Alexandros Doumanoglou, Stylianos Asteriadis, and Dimitrios Zarpalas
\thanks{A. Doumanoglou is with the Information Technologies Institute, Centre for Research and Technology HELLAS, Thessaloniki, Greece and the Department of Advanced Computing Sciences, University of Maastricht, Maastricht, Netherlands (e-mail: aldoum@iti.gr).}
\thanks{S. Asteriadis was with the Department of Advanced Computing Sciences, University of Maastricht, Maastricht, Netherlands (e-mail: stelios.asteriadis@maastrichtuniversity.nl). 
}
\thanks{D. Zarpalas is with the Information Technologies Institute, Centre for Research and Technology HELLAS, Thessaloniki, Greece (e-mail: zarpalas@iti.gr).}
}

\markboth{}
{A. Doumanoglou \MakeLowercase{\textit{et al.}}: UIBE}

\maketitle
\begin{abstract}

An important line of research attempts to explain CNN image classifier predictions and intermediate layer representations in terms of human-understandable concepts. Previous work supports that deep representations are linearly separable with respect to their concept label, implying that the feature space has directions where intermediate representations may be projected onto, to become more understandable. These directions are called interpretable, and when considered as a set, they may form an interpretable feature space basis. Compared to previous top-down probing approaches which use concept annotations to identify the interpretable directions one at a time, in this work, we take a bottom-up approach, identifying the directions from the structure of the feature space, collectively, without relying on supervision from concept labels. Instead, we learn the directions by optimizing for a sparsity property that holds for any interpretable basis. We experiment with existing popular CNNs and demonstrate the effectiveness of our method in extracting an interpretable basis across network architectures and training datasets. We make extensions to existing basis interpretability metrics and show that intermediate layer representations become more interpretable when transformed with the extracted bases. Finally, we compare the bases extracted with our method with the bases derived with supervision and find that, in one aspect, unsupervised basis extraction has a strength that constitutes a limitation of learning the basis with supervision, and we provide potential directions for future research.

\end{abstract}

\begin{IEEEImpStatement}
CNN image classifiers have demonstrated outstanding performance in real-world tasks. They can be used in robotics, visual understanding, automatic risk assessment, and more. However, to a human expert, CNNs are often black-boxes and the reasoning behind their predictions can be unclear.  Recent advances in explainable and interpretable artificial intelligence (XAI and IAI) attempt to shed light on this process. In an attempt to understand intermediate layer representations, one can project them onto a feature space basis that quantifies the presence of different concepts in the representation. This basis is called interpretable because it can make representations more understandable. In the typical approach, constructing an interpretable basis requires access to annotations. This work proposes a novel unsupervised method to learn such a basis, without the need for explicit labels. This can ease the process of obtaining explanations, eliminate annotation costs, save time, and eventually help humans debug and trust deep models.

\end{IEEEImpStatement}


\vspace{5cm}
\section{Introduction}
\label{sec:introduction}
\begin{figure}[t]
    \centering
    \includegraphics[width=3.2in]{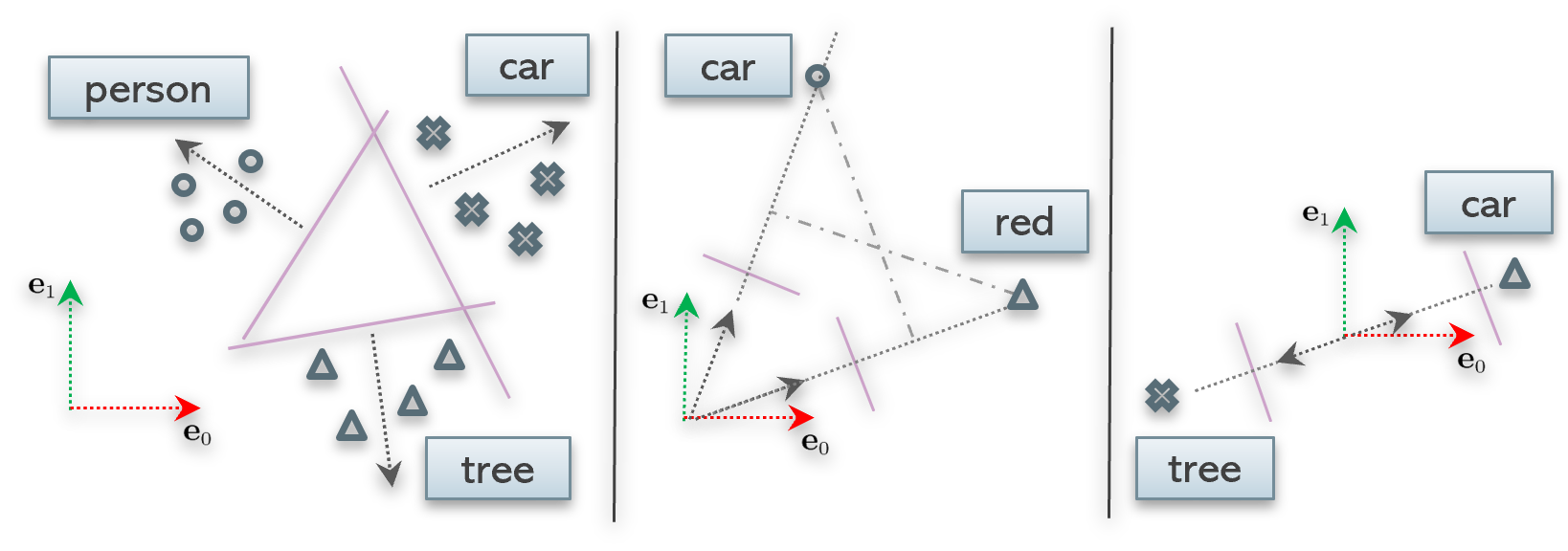}
    \caption{ The natural basis of the feature space is given by $\mathbf{e_0}, \mathbf{e_1}$. \textbf{Left:} An interpretable direction is the direction of the feature space along which, the feature representations of a concept lie. \textbf{Middle:} A case where the \textcolor{purple}{hyperplane} normals of two concept detectors (\textit{car}, \textit{red}) are not orthogonal. In this case, the feature representation of a \textit{car} is also classified as \textit{red} and vice versa. Consequently, $\textit{car}$ and \textit{red} are positively correlated and not (linearly) disentangled. \textbf{Right:} For a pair of mutually-exclusive concepts, the \textcolor{purple}{hyperplane} normals of the two concept detectors may form an angle greater than $90^\circ$. However, in a large dimensional feature space with several detectors of mutually-exclusive concepts, the maximum angle between all pairs of hyperplane normals, is approximately $90^\circ$.}
    \label{fig:concept-directions}
\end{figure}
\begin{figure*}[t]
    \centering
    \includegraphics[width=7.1in]{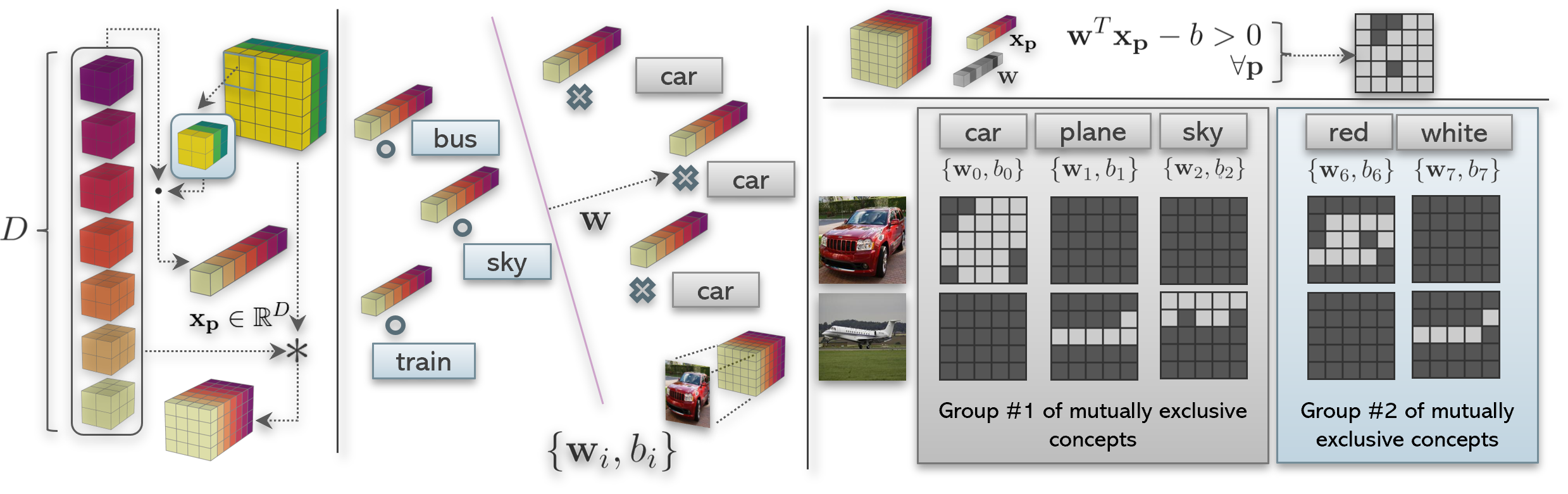}
    \caption{\textbf{Left:} In a standard convolution layer with $D$ filters, all the filters work together to transform each input patch to a feature vector of spatial dimensionality $1 \times 1$. Each spatial element $\p$ of the transformed representation, is assigned a feature vector $\xp \in \R^D$ which lies in the co-domain of the transformation function. Thus the dimensionality of the feature space equals the number of filters in the layer, and each spatial element of the transformed representation, constitutes a sample of this feature space. \textbf{Middle:} To find an interpretable basis for the aforementioned feature space in a supervised way, it means to train a set of linear classifiers (concept detectors), one for each interpretable concept, to separate feature vectors corresponding to image patches with and without the concept. \textbf{Right:} In case classifier training succeeds, the application of the classifier rule ($\w^T \xp - b > 0$) to each spatial element of the representation $\xp$, produces a binary mask which is active for pixels corresponding to image patches containing the concept. We observe, that in a successfully learned interpretable basis, a single pixel is classified positively by at most one classifier, among a group of classifiers that are trained to detect mutually-exclusive concepts.} 
    \label{fig:motivation}
\end{figure*}

\IEEEPARstart{D}{espite} the impressive performance of convolutional neural networks (CNNs) in computer vision image classification tasks \cite{NIPS2012_Krizhevsky_AlexNet,simonyan2014VGG,Szegedy_GoogleNet_2014,he2016ResNet}, the understanding of their inner workings still remains a challenge. In an attempt to shed light on the CNN ``black-box'', the scientific community tries to understand the properties of the intermediate layers' feature space. Early research \cite{szegedy2013intriguing} showed that any possible direction in this feature space may have a semantic meaning, i.e., feature vectors that maximally activate a direction correspond to image patches that share some sort of semantic concept. For instance, image patches of \textit{car doors}, \textit{cat heads} or \textit{people's faces} maximally activate  different directions of this feature space. Beyond this early result, more recently, rigorous experimentation showed that linear separability of features corresponding to different semantic concepts increases towards the top layer \cite{alain2016_Understanding_Linear_Classifiers}. The latter has been attributed to the top layer's linearity and the fact that intermediate layers are enforced to produce representations that are helpful to solve the task at hand.

The fact that linear separation of concept representations is possible (especially for layers near the top) \cite{szegedy2013intriguing}, has motivated attempts in finding feature space directions for specific concepts \cite{kim_TCAV} and constructing an interpretable feature space basis \cite{Zhou_Interpretable_Basis_Decomposition}. In an interpretable basis, each basis vector points towards the direction of a concept's representations. Projecting a representation onto a basis vector quantifies the presence of the respective concept in the representation. An interpretable basis can help to obtain possible explanations regarding the CNN and its predictions. When considering the basis vectors as concept embeddings, an interpretable basis can be used to explain the relationship between concepts and filters, similar to what was proposed in \cite{fong_Net2Vec}. Moreover, it can also be used to interpret predictions of individual examples \cite{Zhou_Interpretable_Basis_Decomposition}, or used to quantify the class sensitivity of the CNN with respect to a concept \cite{kim_TCAV, RTCAV}.

In the typical approach for computing an interpretable basis, the set of interpretable concepts needs to be defined in the form of an annotated concept dataset. Using this dataset, one may have access to labels for intermediate CNN representations. These can be subsequently used to find the orientation of the hyperplane that separates the representations of a concept with respect to representations of other concepts \cite{kim_TCAV}, \cite{Zhou_Interpretable_Basis_Decomposition}. The interpretable basis is constructed by directly using the hyperplane normals as basis vectors.  As with any other supervised approach, using an annotated concept dataset to construct an interpretable basis may increase the fidelity of explanations obtained via that basis. However, this comes at the cost of obtaining the annotations, which is even more prominent when annotations need to be dense (per-pixel) \cite{Zhou_Interpretable_Basis_Decomposition}. Additionally, annotated concept datasets are domain-specific, and thus, explaining CNN classifiers for different domains can become even more costly.

The motivation of this work is based on an observation of how an interpretable basis transforms representations. We explain by examples that projecting a representation onto an interpretable basis and hard thresholding results in a new, sparse, binary representation. Thus, we propose a method that is able to suggest a feature space basis which satisfies this property that holds for interpretable bases. In contrast to the typical approach, the proposed method learns the basis directly from the structure of feature space representations, without requiring access to semantic annotations. In that sense, our method can be considered to be unsupervised. However, without annotations, the final suggested basis vectors are not assigned an explicit concept name. In a real-world setting, the concept name associated with a basis vector could be identified by inspecting samples of image patches whose projected representations onto the vector are maximum. For evaluation purposes though, a procedure to label the basis is required, by assigning a concept label to each basis vector, as in \cite{bau_NetDissection_CVPR} or \cite{mu2020compositional}.

Our work's contributions can be summarized to the following: (i) We present a \textbf{post-hoc}, unsupervised method that suggests an interpretable basis for the feature space of a CNN's intermediate layer. Since post-hoc, the proposed method applies to pre-trained CNN architectures and does not require any form of retraining them. %
(ii) Inspired by related work, we propose simple extensions for two basis interpretability metrics. (iii) We provide a quantitative evaluation of our method on extracting an interpretable basis for the last layer of popular CNNs, demonstrating applicability to standard architectures. We show that our method is able to improve on the interpretability metrics compared to the interpretability of the natural basis \cite{bau_NetDissection_CVPR}, and also compare against a supervised approach \cite{Zhou_Interpretable_Basis_Decomposition} to set a baseline for future works and discuss interesting findings that may help future research.

\section{Background \& Related Work}
\label{sec:related_work}
In this section we discuss background and related work in five areas related to our work. First, we briefly describe prior work on supervised interpretable basis extraction and establish the terminology that is used in this article. Second, we discuss supervised and unsupervised discovering of interpretable feature space directions. Third, we highlight how the proposed method differs from other works proposing sparsity as a measure to build inherently interpretable models. Fourth, we explain the basis labeling problem and potential solutions, and finally, discuss interpretability metrics for assessing the quality of a basis.
\subsection{Supervised interpretable basis extraction and terminology}
As already mentioned in the introduction, each basis vector of an interpretable basis points towards the direction of a concept's representations. To construct an interpretable basis, Zhou et al. \cite{Zhou_Interpretable_Basis_Decomposition} trained a set of binary linear classifiers that separate the CNN's intermediate feature representations based on their semantic meaning. This is accomplished with an densely (per-pixel) annotated concept dataset and implicit use of CNN receptive fields, to assign labels to spatial representation elements of images. Each binary classifier can be considered as a concept detector, since it can separate representations of one concept from representations of other concepts. As already mentioned in Section \ref{sec:introduction}, i) the hyperplane normal directions of the linear classifiers can form a (not necessarily orthogonal, and potentially over/under-complete) basis of the feature space and ii)  projecting a representation onto a basis vector quantifies the presence of the respective concept in the representation. When constructing the basis, each basis vector retains the concept label of its respective concept detector. In a strict sense, the concept detector's bias, which is related to the position of the hyperplane in the feature space, is not part of the basis. For simplicity though, we will retain the association between biases and basis vectors, in such a manner that biases together with the basis vectors form the original concept detectors. For brevity, in this paper, we will use the terms \textit{basis}, \textit{concept detectors}, and \textit{classifiers} almost interchangeably.
\subsection{Discovering concept directions in the CNN feature space} Discovering interpretable directions \cite{szegedy2013intriguing} in the feature space of a CNN image classifier has been previously studied in the literature. In most cases though \cite{alain2016_Understanding_Linear_Classifiers, Zhou_Interpretable_Basis_Decomposition, kim_TCAV, RTCAV}, those directions are directly computed by solving a logistic regression problem that linearly separates CNN's representations based on their concept label. Thus, these methods rely on the existence of an annotated image dataset. To alleviate the need for concept annotations, Ghorbani et al. \cite{ghorbani_ACE} proposed a method to automatically group semantically similar image patches out of an unlabeled image dataset. The image patches of each group could then be treated as samples coming from the same concept. Subsequently, the concept samples may be assigned pseudo-labels and can be used as label-representation pairs to reveal each concept's direction in the feature space of the CNN. The latter may be accomplished via solving the respective logistic regression problem. Ghorbani's approach, even though automated, remains top-down, automating the process of concept speculation. In contrast to \cite{ghorbani_ACE}, our approach is fundamentally different. Our work directly tries to exploit existing structure in the CNN's feature representations, instead of using pseudo-labels to convert the problem to a supervised one.
\subsection{Sparsity in Inherent Interpretable Models} The proposed work shares conceptual similarities with \cite{zhang2018interpretable, liang2020training} and \cite{Losch_Fritz_Schiele_2021}. All previous works are proposing CNN architectures that are inherently interpretable. During training, they enforce intermediate layer representations to be comprised of pixels with sparse activations across feature maps. While we share the same principal idea that sparse pixel activations can lead to more interpretable representations, the proposed method is post-hoc, and has the potential to be applied (possibly) in any pre-trained CNN. In other words, our method suggests a view of the feature space described by the derived basis, that shares similar sparsity properties that other methods enforce during network training. Essentially, and in a more abstract and less strict way, our method reveals the degree that this property is already present in CNNs that were trained without explicitly enforcing this objective.
\subsection{Labeling a feature space basis}
We define \textit{basis labeling} as the procedure of assigning a concept label name to each one of its vectors. When the basis vectors have been learned in a supervised way, the concept label to attribute to each vector is actually known before learning the vector's direction. However, when the basis is learned without annotations (such as the current work) or if the natural feature space basis is considered (as in \cite{bau_NetDissection_CVPR} or \cite{mu2020compositional}), attributing meaning to each basis vector requires putting the vector under test. In the testing procedure, each basis vector is accompanied by a (possibly learned) bias (threshold) to form a linear classifier. Then, for all possible concepts, the suitability of the classifier to separate the representations of one concept (positive samples) with respect to the representations of other concepts (negative samples) is evaluated. Finally, each basis vector is assigned a concept label name based on the evaluation metrics of the aforementioned procedure. It is evident that labeling a basis requires access to a dataset containing concept annotations, such as \cite{bau_NetDissection_CVPR, xiao2018brodenp}, or \cite{cordts2016cityscapes}. Bau et al. \cite{bau_NetDissection_CVPR} assigned one concept label to each vector of the natural feature space basis based on the Broden dataset (which was also introduced in the same work). Later, Mu et al. \cite{mu2020compositional} used the same dataset to label the natural basis with logical compositions of concepts (e.g. the concept of ``blue AND (NOT water)''). In this work, we use \cite{bau_NetDissection_CVPR} to label the bases extracted with our method, while \cite{mu2020compositional} or other potential future works could also be considered.
\subsection{Metrics to evaluate the interpretability of a feature space basis}
In basis evaluation literature \cite{Zhou_Interpretable_Basis_Decomposition}, \cite{bau_NetDissection_CVPR, mu2020compositional}, measuring the interpretability of a basis slightly varies, depending on whether the basis was learned in a supervised way or not. On one hand, in case the basis was learned with supervision, Zhou et al. \cite{Zhou_Interpretable_Basis_Decomposition} used mean average precision (mAP) considering all the classifiers associated with the basis. On the other hand, to assess the interpretability of the natural basis, Bau et al. \cite{bau_NetDissection_CVPR} considered the number of \textit{unique} concept labels that have been assigned to the basis vectors, provided that the performance of the respective classifiers exceeds a threshold. Those labels come from the basis labeling procedure. In \cite{Losch_Fritz_Schiele_2021}, Losch et al. considered Area Under inspectability Curve (AUiC) in order to propose a metric agnostic to a specific threshold. In this paper, we combine ideas from \cite{bau_NetDissection_CVPR} and \cite{Losch_Fritz_Schiele_2021} to propose two metrics that can be used to evaluate the interpretability of a basis.

\section{Motivation}
\label{sec:motivation}

To describe the motivation of our approach, let's assume that we have access to an interpretable basis of a CNN. Let's also assume that the basis was successfully learned, i.e., the CNN representations can be linearly separated based on their semantic label. The latter implies a) the accuracy of the concept detectors is high and b) the CNN has learned to linearly separate (i.e. disentangle \cite{higgins2018disentangledrepresentations}) the aforementioned concepts. In b), disentangled representations can be obtained by projecting representations onto the basis. Inversely, in case the CNN representations could not be linearly separated based on their semantic label, it would mean that the accuracy of concept detectors is low and thus concept disentanglement via a linear transformation is not possible. 

For example, let's consider a basis with five concept detectors, one for each element in the set \{\textit{car, plane, sky, red, white}\}. Consider the images of the red car and white plane of Fig \ref{fig:motivation} - right. If we apply the concept detectors to the intermediate representation of an image patch, we observe that among a group of classifiers detecting mutually exclusive concepts, only one concept detector classifies the patch positively (i.e. as a patch containing the respective concept). For instance, a patch belonging to the concept \textit{car} (such as the one located at second row - second column) is not a \textit{plane} or \textit{sky}, while it is also \textit{red} and not \textit{white}. In that case, \{\textit{car, plane, sky}\} is a group of mutually exclusive concepts, and \{\textit{red, white}\} another one. This simple fact can be summarized to the following observation: \textit{projecting a representation to an interpretable basis and hard thresholding, results into sparse binary representations}.

In this work, we take a non-standard approach to extract an interpretable basis for the feature space of a CNN. Let us consider a set of linear classifiers. In this case, the classification rule dictates projection on the classifier's hyperplane normal and hard-thresholding against the classifier's bias. Based on our previous observation, in case this set of classifiers forms an interpretable basis, applying all the classifiers' rules to a CNN's intermediate representation (with hard thresholding) shall result in a new, transformed representation, which is binary and sparse. By optimizing basis vectors and biases for this sparsity objective, the proposed method is able to suggest an interpretable basis without requiring an annotated concept dataset.

\section{Proposed Method}
\label{sec:proposed_method}
\begin{figure}
    \centering
    \includegraphics[width=3.3in]{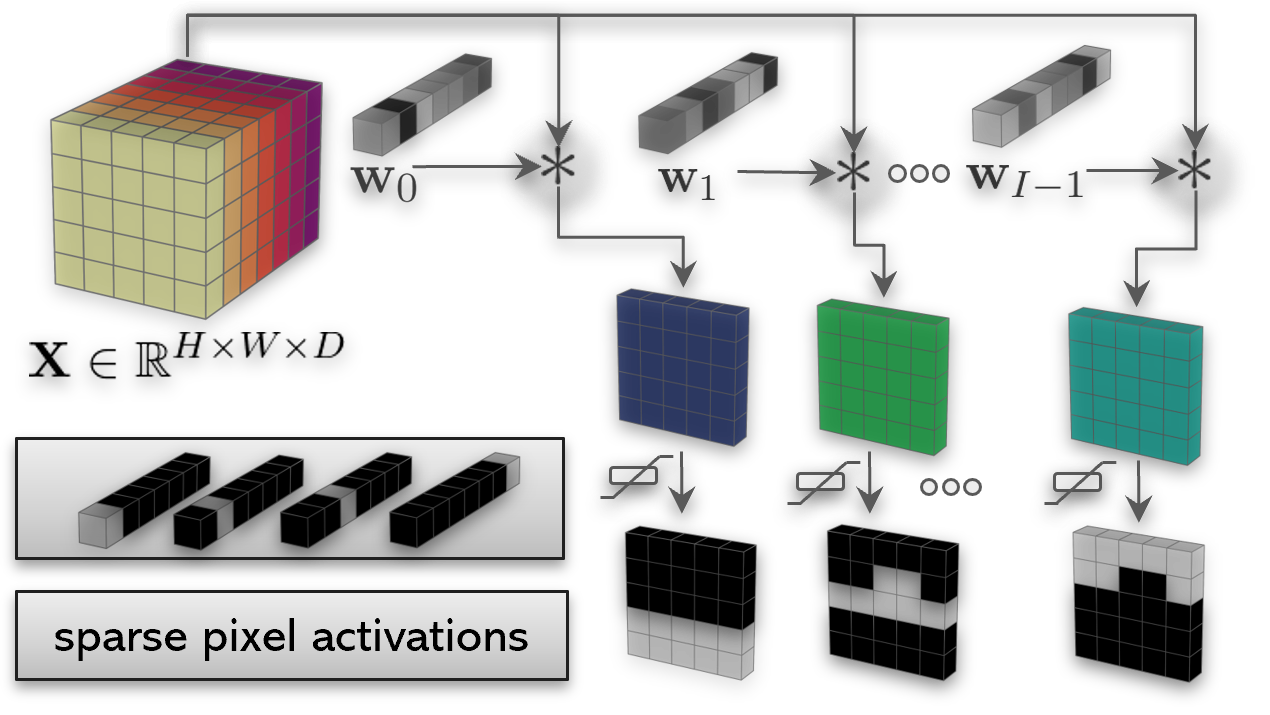}
    \caption{Overview of the proposed method. Without any form of annotation, our method solves for hyperplane normal directions and thresholds of potential binary and linear concept classifiers, driven by the objective that for a single pixel of an image's representation, only a fraction of the classifiers make positive predictions. 
    The application of the linear classification rules to each pixel in the intermediate representation is accomplished by $1\times 1$ convolution between the image representation $\X \in \R^{H\times W \times D}$ and the classifiers' hyperplane normal directions $\wi \in \R^D$, followed by bias subtraction and application of the sigmoid activation function.
    While solving for that objective, the name of the concept that each classifier detects is unknown. In case annotations exist, labeling the basis can be achieved, in a post-processing step, by using methods in related work. In absence of annotations, the concepts can be identified by inspecting samples that the classifiers classify positively.}
    \label{fig:concept}
\end{figure}

\begin{figure}[t]
    \centering
    \includegraphics[width=3.1in]{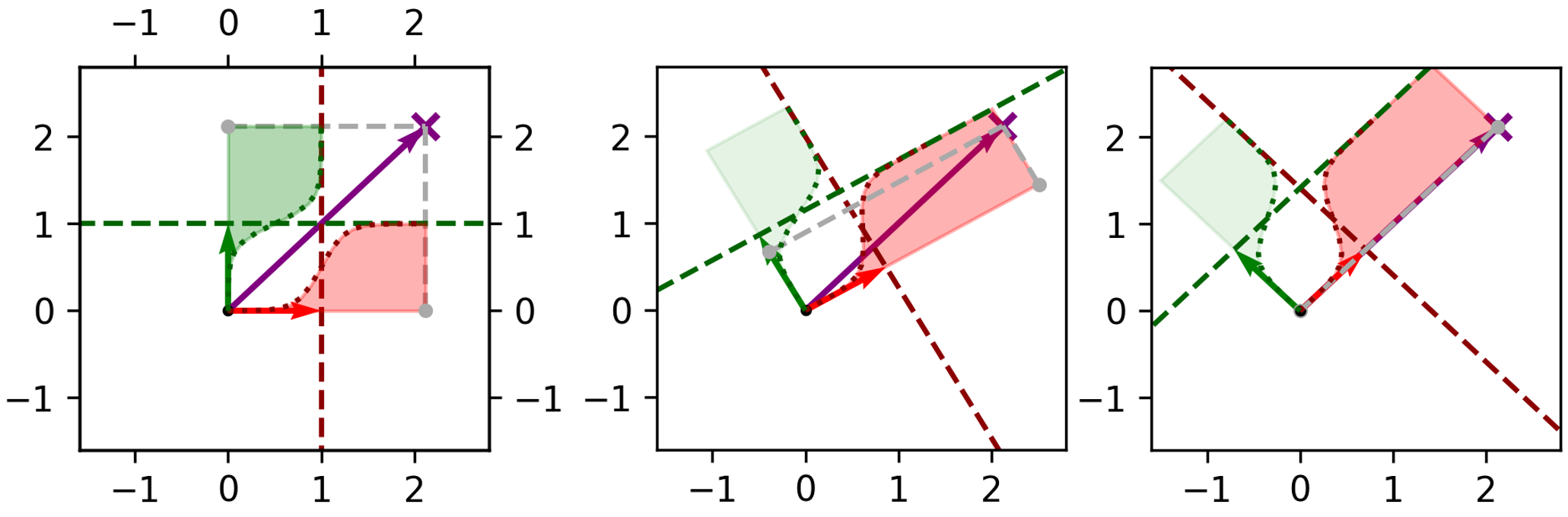}
    \caption{Consider two feature space basis vectors (first -  \textcolor{red}{red}, second - \textcolor{softgreen}{green}) located at the origin (black). Furthermore, consider each feature space basis vector to be accompanied by a bias threshold which together with the basis vector direction constitutes a linear classifier, with the basis vector pointing towards the direction of higher positive classification confidence. On the \textbf{left}, the classifers' separating hyperplanes have been
placed at the location of the bias and indicated by (\textcolor{darkred}{dark red} and \textcolor{darkgreen}{dark green}) dashed lines. Let a feature (\textcolor{purple}{purple}) lie in this space. The projection points of the feature vector on the basis vectors are marked by \textcolor{darkgray}{gray} circular markers. The bottom horizontal and left vertical axes correspond to the standard feature space x and y axis, respectively. The right (top) vertical axis reports the \textcolor{red}{first} (\textcolor{softgreen}{second}) classifier's confidence levels for each point projected in the basis vector's direction. The exact confidence of the classifier at each projected point on its direction, is given by the sigmoid activation function depicted with the dotted (\textcolor{darkred}{dark red}) (\textcolor{darkgreen}{dark green}) curve. On the \textbf{left}, the classifiers attribute the presence of two concepts in the feature, since the projection of the feature on both basis vectors, exceeds the classifiers' biases by a large margin. This is indicated by (\textcolor{darkred}{dark red}, \textcolor{darkgreen}{dark green}) shaded areas under the sigmoid curves. The figure in the \textbf{middle}, depicts the same situation under a rotation of the basis vectors. In that case, the \textcolor{red}{first} classifier makes a confident positive prediction ($\sigma(\cdot) \approx 1.0$) (\textcolor{darkred}{dark red} sigmoid shaded area) for the feature, whereas the \textcolor{softgreen}{second} one makes a confident negative prediction ($\sigma(\cdot) \approx 0.0$) (\textcolor{softgreen}{soft green} sigmoid shaded area). The figure on the \textbf{right} depicts rotation with perfect alignment, where only one of the classifiers classifies the feature positively with high confidence.}
\label{fig:sparsity_rotation}
\end{figure}

In a typical convolutional neural network (CNN) that is trained for image classification, the intermediate layer representations have a cuboid structure. For a convolutional layer, those representations are calculated by applying the same transformation function (a series of dot products equal in number to the number of filters in the layer) to cuboid patches sliced from the representation of the layer beneath. Thus, the dimensionality of the layer's feature space equals the co-domain dimensionality of this transformation function. In this case, this dimensionality is equal to the number of filters in the layer. Feature vectors at different spatial locations of the cuboid, correspond to different samples from this feature space (Fig. \ref{fig:motivation} - left). This treatment of the feature space has been also considered in \cite{bau_NetDissection_CVPR, Zhou_Interpretable_Basis_Decomposition,fong_Net2Vec, mu2020compositional}

Let $D \in \N^+$ denote the dimensionality of a layer's feature space, and $\xp \in \R^D$ an element in this space at the spatial location $\p=(x,y)$ (Fig. \ref{fig:motivation} - left). In a convolutional layer, $D$ equals the layer's total number of hidden units or, as otherwise mentioned, output channels. %
Let's consider a set of $I \in \N^+, I \leq D$ linear classifiers to form a (possibly) interpretable basis. The $i$-th classifier is characterized by its hyper-plane's normal direction $\w_i \in \R^D$ and bias $b_i \in \R$, $i\in \I ,\, \I = \{0,1,...,I-1\}$. Additionally, each one of those classifiers is responsible to quantify the presence of one concept in $\xp$. Last, for the reasons discussed in the Section \ref{sec:orthogonality}, we also consider $\wi^T \wi = 1 \, \forall i,\, \w_i^T\w_j = 0 \, \forall \, i,j : i \neq j$, i.e. $\{\w_i\}$ should form an orthonormal basis. We consider $I$ to be a hyper-parameter of the method and, without loss of generality, when $I<D$, the orthogonal basis can be trivially completed to dimensionality $D$ in order to represent a rotation of the feature space. The additional $D-I$ directions can be considered as a non-interpretable \textit{residual}.

The overall concept of our method is depicted in Fig. \ref{fig:concept}. First, we record CNN intermediate layer representations for images coming from an unlabeled dataset. Starting from the representation of an image $\X$, we project each spatial element $\xp$ onto all the vectors of the basis $\wi$ via $1\times 1$ convolution. This operation transforms each pixel of the image representation to the new basis. The result is a new, transformed, cuboid representation. In the new representation, the pixel $\p$ of the $i$-th feature map has a value equal to the projection of $\xp$ onto $\wi$. Subsequently, we threshold the projections with a learned bias $b_i$ and use a sparsity objective to enforce each pixel to have a sparse thresholded representation across feature maps.

\begin{figure*}[t]
    \centering
    \includegraphics[width=7.1in]{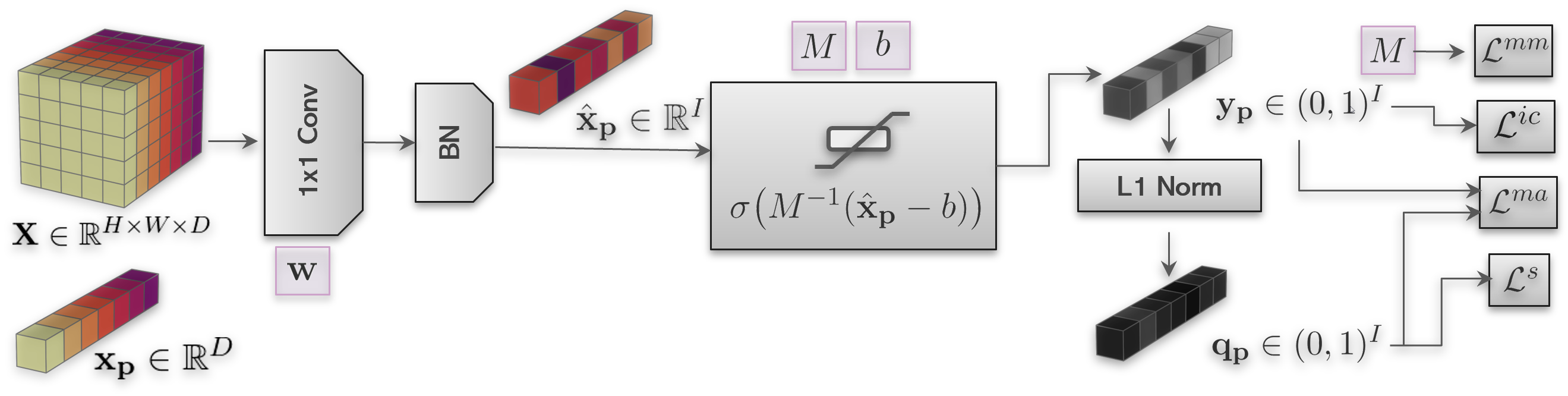}
    \caption{The basis learning pipeline of the proposed method. Learnable parameters are given in purple next to the operations that actually use them.}
    \label{fig:method}
\end{figure*}

To formalize all the previous discussion, consider the standard binary sigmoid classifier $\sig(\wi^T\xp-b_i)$ which, since $||\wi||=1$ and for full expressivity,  requires an additional parameter $M_i \in \R^+$, such that, $\ypi = \sig(\frac{1}{M_i}(\wi^T\xp-b_i))$. $M_i$ is controlling the margin between the abscissas corresponding to the extremas of the sigmoid and $\ypi \in (0,1)$ denotes the confidence of classifier $i$ to classify $\xp$ positively. Without loss of generality and for mathematical and implementation convenience, we standardize the feature space with batch normalization \cite{ioffe2015batch} and without affine parameters. We do so, just after projecting $\xp$ to $\wi$ and before subtracting the bias $b_i$ or dividing by $M_i$. As already mentioned, the projection of each $\xp$ to the new basis is accomplished via standard $1\times 1$ convolution with $D$ input and $I$ output channels. While searching for $\wi$, the standardization of the feature space allows treating the magnitude of projections $\xp^T\wi$, biases $b_i$, and margin coefficients $M_i$ in the same scale, respectively, regardless of $i$. Thus, this allows us to make a simplification to the parameter space of our model and consider $b_i = b$ and $M_i = M$ (i.e. equal biases and margins in the standardized space) for all $i$. Orthogonality of the extracted basis is enforced by using \cite{lezcano2019trivializations}. The learnable parameters of our model are simply $\wi, b$ and $M$, while $b_i$ and $M_i$ can be later recovered by inverting the standardization process. A graphical explanation of the principal idea in the 2D feature space is provided in Fig. \ref{fig:sparsity_rotation} and the pipeline of the proposed method is given in Fig. \ref{fig:method}.

In the rest of the section, we introduce the loss terms that we use to derive an interpretable basis. For notation convenience, we assume $\p$ to vary across the spatial dimensions of all image representations in the dataset. Moreover, a pixel $\p$ is considered to be assigned to the $i$-th concept detector, when for the given $\xp$, $\ypi \gg 0.5$. In that case, we also say that $\xp$ is classified positively by the same concept detector. In a similar analogy, we mention $\xp$ to be classified negatively by the $i$-th concept detector whenever $\ypi \ll 0.5$.

\noindent\textbf{Sparsity Loss (SL)} Let $\yp = [\ypk{0},\ypk{1},...,\ypk{I-1}]^T$ denote the vector of activations containing the classification results for $\xp, \forall i \in \I$. The criterion that guides our search for $\wi$ implies sparsity in this vector of activations. Under the sparsity criterion, each pixel $\p$ is classified positively only by a portion of the classifiers in the new basis. We use entropy as a sparsity measure and define the sparsity loss $\loss^{s}$ as:
\begin{equation}
\label{eq:sparsity_loss}
\loss^{s} = \averagep\Big[-\sum_{i\in \I}\qpi\logtwo(\qpi)\Big]
\end{equation}
with
\begin{equation}
\label{eq:relative_scaling}
\qpi = \frac{\ypi}{\sum_{i \ \in \I}\ypk{i}}
\end{equation}

\begin{figure}[t]
    \centering
    \includegraphics[width=3.4in]{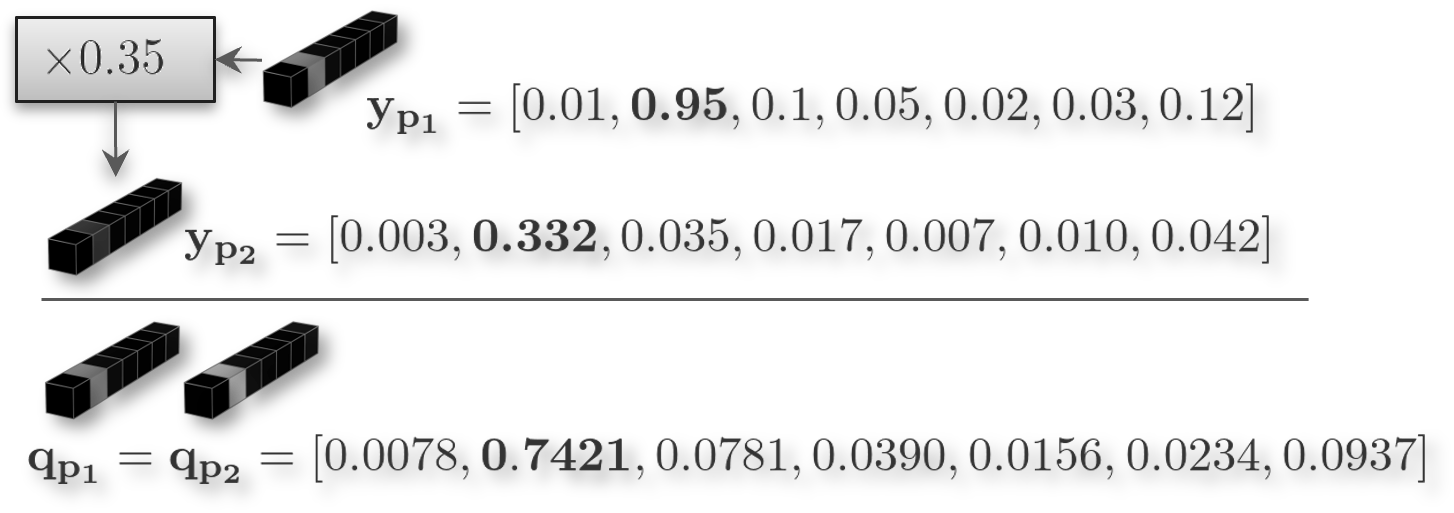}
    \caption{An example why pixel activation sparsity, which is enforced through entropy, is not alone sufficient to provide a meaningful basis. Since entropy can be applied only on probability distributions, $\yp$ is L1 normalized to $\qp$ before enforcing sparsity. This may lead to a set of classifiers that satisfy the sparsity criteria on $\qp$ but actually none of the classifiers classifies $\xp$ positively (i.e. with high confidence $\gg 0.5$). Thus, $\xp$ has no concept assigned to it. If we exaggerate to many $\p$, this may lead to a basis that does not classify positively any of the pixels. In the figure, $\yp_2$ is derived by scaling $\yp_1$ by 0.35. While both $\p_1$, $\p_2$ have sparse activation in the probability scale (described by $\qp$), only $\p_1$ has a concept assigned to it with high confidence. To mitigate this, we introduce the maximum activation loss which enforces strong activation magnitudes from the most positively confident classifiers.}
    \label{fig:sparsity_not_enough}
\end{figure}

\noindent\textbf{Maximum Activation Loss (MAL)} The sparsity criterion alone is not sufficient to extract a meaningful basis. This is better understood when considering the fact that entropy is applied in a relative scaling of the activation magnitudes, due to 
(\ref{eq:relative_scaling}). Thus, when optimizing $\loss^{s}$
alone, a pixel's activations may be considered sparse by eq (\ref{eq:sparsity_loss}), but $\xp$ might still be classified negatively by all concept detectors in the basis, i.e. $\mathop{\max}_{i} \ypi < 0.5$. For a meaningful basis, we would like to have each pixel assigned to at least one concept detector. To this end, we add an additional loss term $\loss^{ma}$ that encourages the most confident concept detector to not only be the most confident in a relative scale (compared to other concept detectors) but also in an absolute scale, reporting high confidence levels towards $1$ (Fig. \ref{fig:sparsity_not_enough})
\begin{equation}
\label{eq:max_act_loss}
\loss^{ma} = \averagep\Big[-\sum_{i \in \I} \qpi \logtwo(\ypi)\Big]
\end{equation}
In (\ref{eq:max_act_loss}), $\logtwo$ is chosen for its strong guiding gradient when $\mathop{\max}_{i} \ypi \ll 0.5$. From another viewpoint, this loss in combination with the sparsity loss, imposes each pixel to be classified positively with high confidence from the most confident classifiers and negative with high confidence from the remaining classifiers.

\noindent\textbf{Inactive Classifier Loss (ICL)} The two previous losses while they encourage assigning each pixel to a basis vector, they do not encourage, in any way, diversity in the assignments. For instance, all pixels could be assigned to one concept detector, with the rest of the detectors having no pixels assigned to them. In that case, the classifiers associated with an empty pixel set (i.e. when no pixel is assigned to them), actually never classify a pixel positively and thus the sparsity criterion can be easier fulfilled. Besides, if all pixels in the dataset are classified negatively by a classifier, then this classifier does not convey any meaningful information, it cannot serve as a concept detector and is redundant. 

To moderate this issue we introduce the inactive classifier loss. We design a loss term that linearly penalizes basis vectors with a few number of pixel assignments. This number is defined as a percentage threshold over the total number of pixels in the dataset. Instead of specifying this threshold for each $i \in \I$ individually, we introduce a set of hyper-parameters to make this more manageable. Let $\am \in [0,1]$ denote a percentage coefficient with $\sum_{\mu} \am = 1$ and $\alpha_{0} \ge \alpha_{1} \ge ... \ge \alpha_{N-1}$, $N \in \N^+$, $\mu = \{0,1,..., N-1\}$. We split $\I$ in $N$ partitions with each partition having $n_{\mu} \in \N$ elements:
\begin{equation}
\nm =  
\begin{cases}
    \floor{\am I} + 1 & \text{$\mu >= I - R$} \\
    \floor{\am I} & \text{otherwise}
\end{cases}
\end{equation}
with $R = I - \sum_{\mu} \floor{\am I}$, and $\floor{\cdot}$ denoting the floor operation. The previous procedure ensures that $\sum_{\mu}\nm = I$ while $\nm$ remains integer. Let $\tau \in [0,1]$ denote a  percentage threshold over the total number of pixels in the dataset. We distribute $\tau$ across the concept detectors using a weighting scheme that utilizes the same weight $\omega_{\mu} \in \R^+$ for all detectors in the same partition. The $i$-th concept detector is penalized whenever the percentage of pixels assigned to it falls below the threshold $\nu_{i}$ given below:
\begin{equation}
    \label{eq:inactive_loss_target_thres}
    \nu_{i} = \frac{\omega_{\mu}\tau}{\sum_{\mu}\omega_{\mu}\nm}
\end{equation}
From (\ref{eq:inactive_loss_target_thres}) it becomes apparent that all concept detectors in the same partition share the same threshold. Finally, we define the \textit{inactive classifier} loss as
\begin{equation}
\loss^{ic} = \averagei\Big[\frac{1}{\nu_{i}}\text{ReLU}\big(\nu_{i}-\averagep[{\ypk{i}}^\gamma)]\big)\Big]
\end{equation}
where the factor $\frac{1}{\nu_{i}}$ before the rectified linear unit activation function ($\text{ReLU}$) \cite{DLBook} normalizes the loss to be 1 when all concept detectors classify negatively the whole pixel dataset. The exponent $\gamma \in \R^+, \gamma > 1$, acts as a sharpening operator on $\ypk{i}$, in order to attenuate non-confident predictions that lie around 0.5.

\noindent\textbf{Maximum Margin Loss (MML)}
Since $M$ controls the margin between the abscissas corresponding to the extremas of the sigmoid classifier, we add an additional loss term that encourages a large classification margin in a similar sense as in the Support Vector Machine \cite{SVM}. We enforce $M$ to be a positive scalar via the parameterization $M = 1/t^2$ and simply define the maximum margin loss as:
\begin{equation}
    \loss^{mm} = \frac{1}{M} = t^2
\end{equation}

Conclusively, we introduce four loss terms that guides search for an interpretable basis. First, the Sparsity Loss (SL) which enforces each pixel to be classified positively by only a fraction of the concept detectors in the basis. Second, the Maximum Activation Loss (MML) which in combination with the sparsity loss enforces the most confident predictions in the relative scale (as implied by $\qp$), to also be confident in an absolute scale (as given by $\yp$ and close to $1$). Third, the Inactive Classifier Loss (ICL), which penalizes classifiers that never classify any pixel positively and last, the Maximum Margin Loss (MML) which enforces large hyperplane separation margin (in the SVM sense) between the positive and negative predictions of the classifiers.

\section{Basis orthogonality}
\label{sec:orthogonality}
To explain why we apply orthogonal constraints for extracting an interpretable basis, it is better to individually consider cases where those contraints are absent. To begin with, let's consider the case where two concepts in a concept pair belong to different groups of mutually exclusive concepts. In a slightly informal way where strict linear relation is not considered, this makes the two concepts either independent from each other, or positively correlated, since mutual-exclusivity implies negative correlation. In the first case, it is apparent that the respective basis vectors should be orthogonal. To give a counter example, let's consider the concept \textit{car} from the group of \textit{objects} = \{\textit{car, tree, person}\} and the concept \textit{red} from the group of \textit{colors} = \{\textit{red, green, blue}\}. In case the angle between the basis vectors of these two concepts is less (greater) than $90^{\circ}$, some feature vectors that are classified as \textit{car} will inevitably also be classified as (not) \textit{red} and vice versa (Fig. \ref{fig:concept-directions} - Middle). This relation implies dependence which is contradictory to our initial assumption that the two concepts are independent. While this bias may be encoded in the CNN's weights, this fact also means that the two concepts are not (linearly) disentangled, eventually harming the interpretability of the feature space. Since our primary goal is to search for an interpretable basis, given the previous discussion, we know a-priory that a non-orthogonal basis cannot satisfy the interpretability criteria for independent concepts. 

For the second case, where the two concepts are positively correlated, the two concepts could possibly be related with a \textit{has-a} relationship. For instance, \textit{car} has a \textit{car-door} and a \textit{car-wheel}. In this case, an image patch of the concept \textit{car-door} or \textit{car-wheel} may also be classified as \textit{car}. Vice versa, a representation of a \textit{car} may have positive components in the direction of \textit{car-door} and \textit{car-wheel}, to justify the \textit{has-a} relationship. This case is not handled by the proposed method. However, the primitive concepts, \textit{car-door} and \textit{car-wheel}, are mutually exclusive. 

\begin{figure}
    \centering
    \includegraphics[width=3.3in]{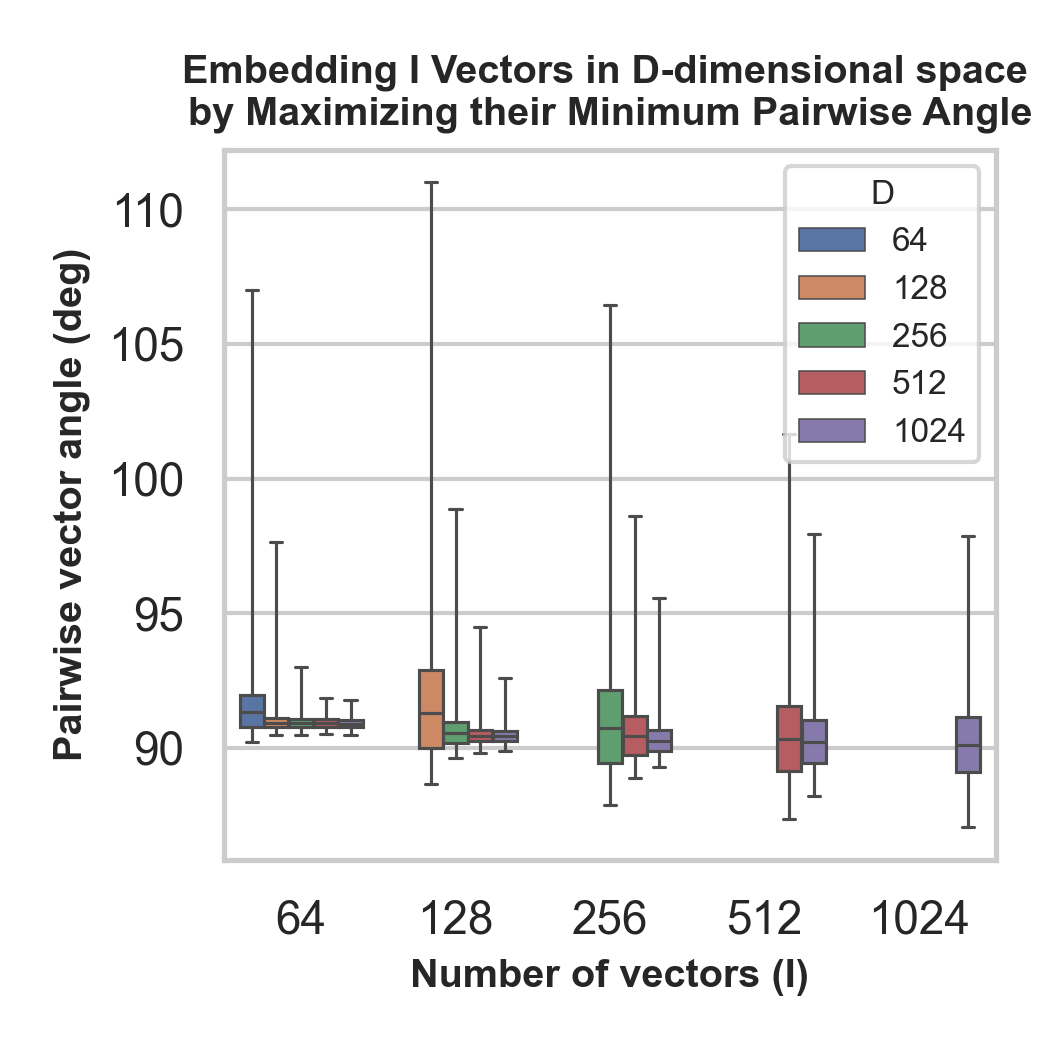}
    \caption{Pairwise vector angle distribution when solving the tammes problem. Extremas of the error bars correspond to the minimum and maximum vector pair angle. The horizontal line in the box is equals to the mean of the distribution and box widths are equal to the standard deviation.}
    \label{fig:tammes}
\end{figure}

Thus, for this last case, considering two concepts coming from the same group of mutually-exclusive concepts, it could be reasonable to expect that this mutual exclusivity, which implies negative correlation, is also encoded in the angle between the respective basis vectors. In that case, the angle between the respective basis vectors could be greater than $90^{\circ}$ Fig. \ref{fig:concept-directions} - Right. To investigate the degree that this is possible, we formulate the problem in a way that is independent from input data. To construct an (ideal) basis for negatively correlated concepts, one might consider embedding $I$ concept vectors in a $D$ dimensional space by maximizing the minimum angle across all pairs of vectors. As it turns out this is linked to spherical coding theory \cite{whyte1952uniqueSphereCode} and the tammes problem \cite{tammes1930origin}. Although more sophisticated approaches exist \cite{kottwitz1991densestSphericalCode, Wang2009SphericalCodes}, we tried to approximately solve the tammes problem via directly maximizing the minimum pairwise vector angle with gradient decent. Experimental results showed that the resulting embedding vectors, in cases where $I \ge 64$, are close to orthogonal. Fig \ref{fig:tammes} depicts distribution statistics for various pairs of $I,D$ with $I \leq D$. Conclusively, we argue that an orthogonal basis can cover (under some approximation) independent and mutually exclusive concepts but not concepts that are positively correlated.

\section{Evaluation Metrics}
\label{sec:evaluation_metrics}
\subsection{Basis labeling and classifier validation scores}
To quantitatively evaluate the bases extracted with our method for interpretability, we use a two step process. First, after deriving a basis, we use %
the work of Bau at al. \cite{bau_NetDissection_CVPR, zhou_NetDissection_PAMI}, to assign a concept label to each classifier associated with the basis vectors. Let $\phi(i,c,\K) \in [0,1]$ denote a metric score function that is used to measure the \textit{suitability} of the classifier $i$ ($\{\w_i,b_i\}$) to accurately detect concept $c$ in the annotated concept dataset $\K$. The concept label that is assigned to classifier $i$ is the one that maximizes $\phi(i,c,\Ktr)$ across $c$ over the training split $\Ktr$ of the concept dataset. Subsequently, in the second step, and using the validation split of the concept dataset $\Kvl$, each classifier is assigned a validation score $\phi(i,\cstar_i,\Kvl)$, with $\cstar_i$ denoting the concept label assigned to the classifier during the first step.  For the choice of $\phi$ we use Intersection Over Union (IoU), as originally proposed in \cite{bau_NetDissection_CVPR} and also used in \cite{mu2020compositional,fong_Net2Vec}:
\begin{equation}
    \phi(i,c,\K) = \frac{\sum_{\k \in \K} |\M^i(\k) \cap \L^c(\k)|}{\sum_{\k \in \K} |\M^i(\k)\cup \L^c(\k)|}
    \label{eq:iou}
\end{equation}
In (\ref{eq:iou}), $\M^i(\k)$ denotes the upsampled, hard-thresholded (binarized) map of image $\k$. $\M^i(\k)$ is obtained by applying the rule of the $i$-th classifier ($\wi^T\xp - b_i > 0$) to each $\xp$ of the image's representation followed by upsampling to the resolution of the original image. Moreover, $\L^c(\k)$ denotes the ground truth segmentation map of image $\k$ for concept $c$ and $|\cdot|$ denotes the cardinality of a set. %
Overall, to label the bases and compute classifier validation scores, we use the exact scheme of \cite{bau_NetDissection_CVPR} with two differences. First, we consider a train/test split of the concept dataset as originally proposed in \cite{fong_Net2Vec} and second, for hard-thresholding in $\M^i(\k)$, we use the biases learned from our method, instead of using the statistical quantile learning of \cite{bau_NetDissection_CVPR}.

\subsection{Overall basis interpretability scores}
Inspired from \cite{bau_NetDissection_CVPR} and \cite{Losch_Fritz_Schiele_2021} we propose two metrics $\Score1$ and $\Score2$ that can be used to measure the interpretability of a basis. Those metrics, essentially aggregate the aforementioned individual classifier validation scores into scalar values that can summarize the interpretability of a basis. 

The first, counts the number of concept detectors in the basis with a validation score better than a threshold $\xi$. In order to make it threshold agnostic, we measure the area under the indicator function ($\mathbbm{1}(x)$) for all $\xi \in [0,1]$:
\begin{equation}
\label{eq:score1}
\Score{1} = \int_{0}^{1} \sum_{i=0}^{I-1} \mathbbm{1}_{x\ge \xi}\big(\phi(i,c^*_i,\Kvl)\big)d\xi
\end{equation}
\noindent This metric is similar to what was proposed in \cite{Losch_Fritz_Schiele_2021} with two differences. First, we use IoU as the choice of $\phi$ in order to comply with our intention to use \cite{bau_NetDissection_CVPR} for labeling the basis. And second, unlike \cite{Losch_Fritz_Schiele_2021},  we do not normalize (\ref{eq:score1}) with the number of vectors in the basis, in order to be able to make absolute comparisons between scores for bases of different sizes. 

The second metric, counts the number of unique concept labels over the set of labels whose respective concept detectors exhibit performance better than $\xi$. This metric is the same as the one proposed in \cite{bau_NetDissection_CVPR}. Inspired by \cite{Losch_Fritz_Schiele_2021}, and with the intention to also make it agnostic to the threshold $\xi$, we use the area under curve:
\begin{equation}
\label{eq:score2}
\Score{2} = \int_{0}^{1} \psi(\xi)d\xi
\end{equation}
with $\psi(\xi) = |\{\cstar_i \,| \, \exists \, i: \phi(i,\cstar,\Kvl) \ge \xi\}|$, i.e. the number of unique concept detectors exhibiting performance better than $\xi$.

\section{Experimental Results}
\label{sec:experimental_results}
\begin{figure}
    \centering
    \includegraphics[width=3.4in]{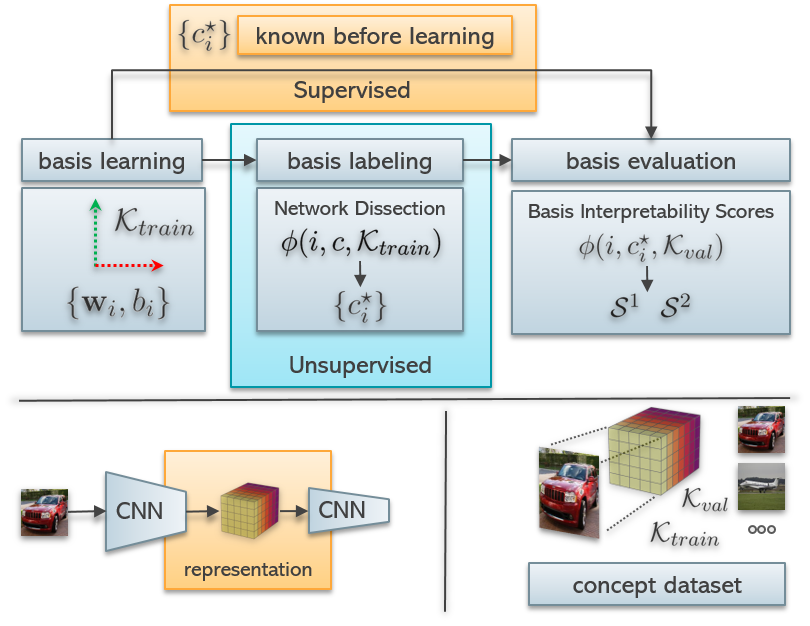}
    \caption{The pipeline for evaluating the interpretability of a basis. The basis labeling procedure is only required when the learned basis was derived in an unsupervised way or when considering the natural feature space basis. In the supervised case, the concept label for each basis vector is actually known before learning the respective concept detector.}
    \label{fig:evaluation}
\end{figure}
\textbf{Overall Evaluation Approach} 
To the best of our knowledge, the proposed method is the first unsupervised method to suggest an interpretable basis. In addition to this, and once again to the best of our knowledge, except from \cite{bau_NetDissection_CVPR} which performs this in a statistical manner, the proposed method is the first unsupervised method to also provide an estimate for the position of the hyperplane that separates each concept's representations from the representations of other concepts. Therefore, we quantitatively evaluate the interpretability of the bases extracted with the proposed method against the interpretability of the natural feature space basis (\textit{baseline}). Apart from this, we quantitatively evaluate the bases extracted with our method with the bases extracted via the supervised approach of \cite{Zhou_Interpretable_Basis_Decomposition} and thus setting a baseline for future unsupervised works.

An exhaustive search and ablation study over all hyper-parameters is difficult, due to the sheer number of parameters, combinations and computational resource constraints. Nevertheless, the results presented below show that by making simple and intuitive hyper-parameter choices, one may obtain a basis that is more interpretable than the natural. In \cite{bau_NetDissection_CVPR}, Bau et. al proved experimentally that the natural feature space basis is more interpretable than other random bases. In this work, we build on the previous findings of \cite{bau_NetDissection_CVPR} and show that the proposed method is able to suggest a basis which is more interpretable than the natural and consequently more interpretable than most other random bases. In short, the main advantage of the proposed method is that it can provide an improvement over the interpretability of the natural basis, and do so, without annotations. Moreover, future, more exhaustive work on fine-tuning strategies has the potential to further improve interpretability. 

In all of our experiments we used the Broden \cite{bau_NetDissection_CVPR} concept dataset to probe the networks and obtain intermediate layer feature representations. Except for comparison with the supervised approach (Section \ref{sec:experimental_supervised}), where we only used the \textit{object} and \textit{part} categories of the dataset, on all other experiments we used the complete set of concept categories, namely \{\textit{scene, object, part, texture, material, color}\}. In all experiments, we used post ReLU activations of the considered network's \textit{last-layer}. A network's \textit{last-layer} refers to the latest convolutional or max-pooling layer where the representation remains spatial, before the flattening to the latest fully-connected one. 

To learn an interpretable basis with the proposed method, we used the training split of the concept dataset. Next, we used the same training split to label the basis using \cite{bau_NetDissection_CVPR}, and finally, we calculated the basis interpretability scores (eq. (\ref{eq:score1}), (\ref{eq:score2})) using the validation split of the same dataset. Annotation labels were only used to label the bases and perform quantitative evaluation, and were not used in any way to learn the aforementioned bases. Regarding the evaluation of the natural basis (\textit{baseline}), we used $\w_i = \e_i$, $\e_i = [\underbrace{0,...,0}_{i \text{ times}},1,\underbrace{0,...,0}_{D-i-1 \, \text{times}}]^T$ and we chose the thresholds $b_i$ according to the top 0.005 -- quantile among the population of projected representations, as suggested by \cite{bau_NetDissection_CVPR}. The rest of the evaluation pipeline was the same as before. Finally, to establish comparisons, we also used the same interpretability score functions of Section \ref{sec:evaluation_metrics}, in order to evaluate the bases extracted with the supervised approach of \cite{Zhou_Interpretable_Basis_Decomposition}. In that case, the bases were learned in a supervised way using the training split of the concept dataset. Given the a-priory known concept labels of the basis vectors, evaluation was performed on the validation split of the dataset, ommiting the basis labeling procedure which is not required. The overall evaluation pipeline is depicted in Fig. \ref{fig:evaluation}. 

\textbf {Basis Learning Details} To learn each one of the basis, we used the Adam \cite{kingma2014adam} optimizer with the default beta parameters ($0.9, \, 0.999$) provided by the PyTorch \cite{Pytorch} implementation. We fixed the learning rate to $0.001$ and did not employ any form of learning rate scheduling. In all cases, basis learning lasted for $300$ epochs. Batch size was a variable that varied across our experiments and its value was based solely on the available GPU memory resources. The values we used, approximately lied in the interval $\approx [800 - 3600]$.

\textbf{Hyper-parameters} We kept most of the hyper-parameters of our method fixed to the same values across all the presented experiments, except for the parameters we wanted to ablate. We linearly combined the loss terms with the weights given in Table \ref{tab:weights}. Empirical evaluation showed that $\lambda^{ma}$ should have higher weight than $\lambda^{s}$ due to the fact that even if the entropy sparsity criterion is fulfilled, the basis may still be not meaningful (Fig. \ref{fig:sparsity_not_enough}). The choice for the rest of the weights was guided by intuition for the  relative importance across loss terms. In all of our experiments we used $I=D$, while extensive study for cases where $I<D$ is left for future work.

\textbf{Parameter Initialization} In all of our experiments we initialize the basis vectors with the vectors of the natural feature space basis (i.e $\wi=\e_i$). We also initialize $t$ and $b$ with $t=0.5$ and $b=0.5$.

\begin{table}[]
\centering
\caption{The loss weight coefficients that we used for learning all our bases. In case of ablation studies, the deviations from these values are given in the respective Section. The superscript of each weight follows the notation of the respective loss.}
\label{tab:weights}
\resizebox{0.5\columnwidth}{!}{
\begin{tabular}{|c|c|c|c|c|}
\hline
$\lambda^s$ & $\lambda^{ma}$ & $\lambda^{ic}$ & $\lambda^{mm}$ \\ \hline
$2.0$ & 5.0 & 5.0 & 0.5 \\ \hline
\end{tabular}
}
\end{table}

\subsection{Ablation studies}
In this section, we present three ablation studies regarding Maximum Margin (\textit{MML}) and Inactive Classifier (\textit{ICL}) losses. To do so, we choose two different CNN architectures trained on two different datasets. In particular, we extract bases for the \textit{last layer} of ResNet18 \cite{he2016ResNet} and VGG16 \cite{simonyan2014VGG} with batch normalization blocks (VGG16BN). The ResNet18 that we used, was trained on Places365 \cite{zhou2017places}, while VGG16BN was trained on ImageNet \cite{deng2009imagenet}.

\begin{figure}
    \centering
    \includegraphics[width=3.4in]{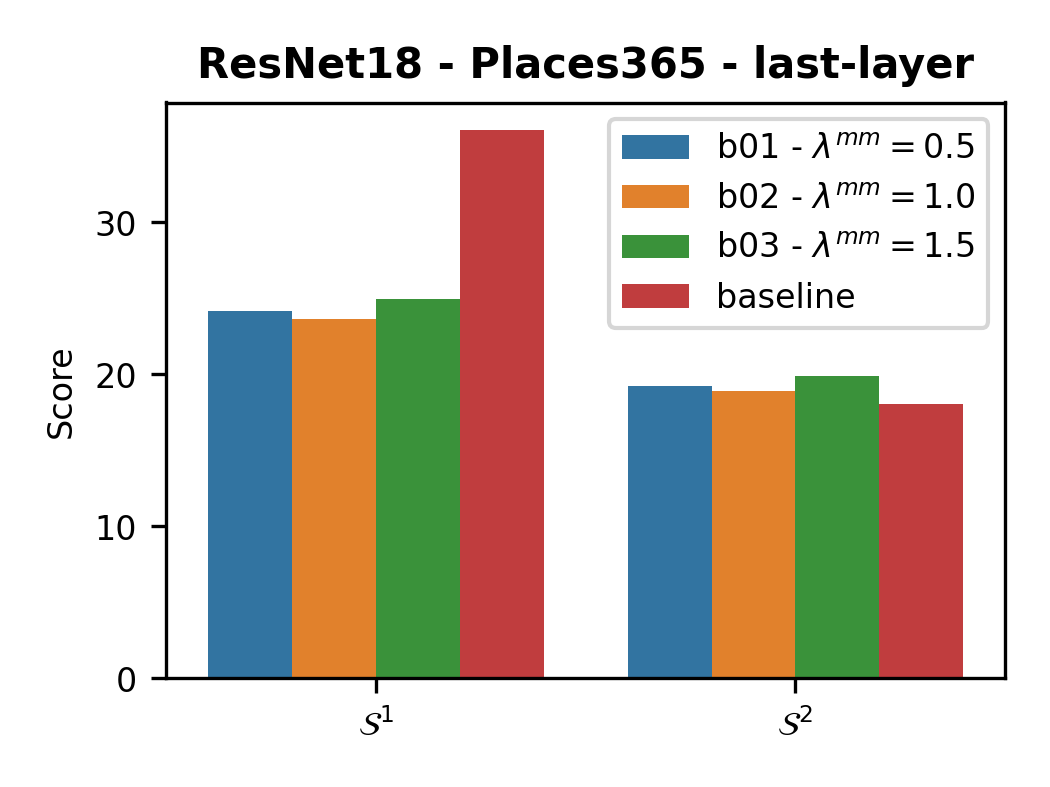}
    \caption{Ablation study for $\lambda^{mm}$. ICL is not used in these experiments. Without ICL, the interpretability of the extracted basis is significantly worse than the baseline in terms of $\Score1$, and slightly better than the baseline in terms of $\Score2$.}
    \label{fig:resnet_ablation_margin}
\end{figure}

\begin{figure}
    \centering
    \includegraphics[width=3.4in]{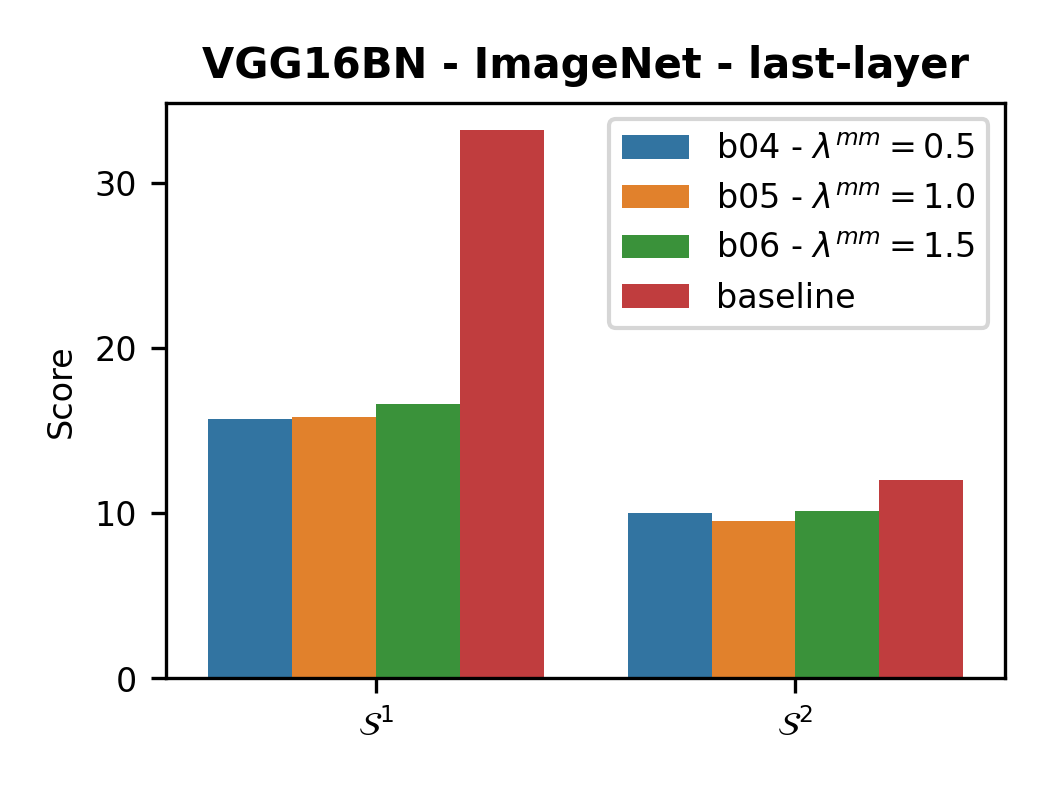}
    \caption{Ablation study for $\lambda^{mm}$. ICL is not used in these experiments. Without ICL, the interpretability of the extracted basis is worse than the baseline in terms of both $\Score1$ and $\Score2$.}
    \label{fig:vgg16bn_ablation_margin}
\end{figure}

\textbf{Ablation of MML weight in absence of ICL}
In the first ablation, we set $\lambda^{ic} = 0$ (i.e. completely eliminated the Inactive Classifier Loss) and varied $\lambda^{mm}$ to take values from the set $\{0.5, 1.0, 1.5\}$. The basis interpretability scores are given in Fig. \ref{fig:resnet_ablation_margin} and \ref{fig:vgg16bn_ablation_margin}. For both networks, we observe that all the bases extracted with the proposed method score significantly lower in terms of $\Score1$ than the baseline. In absence of ICL, this fact was actually expected, for the reasons described in Section \ref{sec:proposed_method}. For ResNet18, the proposed method extracted bases that were slightly more interpretable than the baseline according to $\Score2$, while for VGG16BN and the same metric, none of the learned basis scored higher compared to the baseline. Overall, we could say that for those cases, the sensitivity of the method with respect to $\lambda^{mm}$ was rather small. This stems from the fact that, according to those metrics, the learned bases are approximately equally interpretable, even though they were learned using different $\lambda^{mm}$.


\begin{figure}
    \centering
    \includegraphics[width=3.4in]{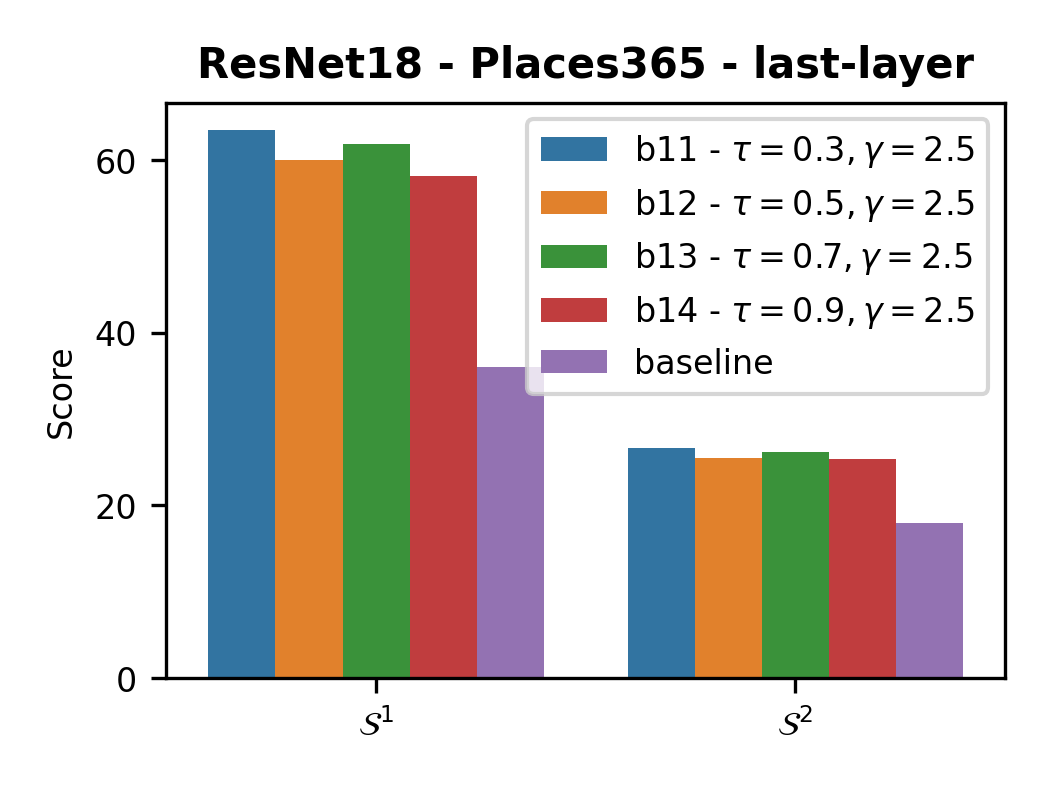}
    \caption{Ablation with respect to $\tau$. With the addition of ICL the interpretability of the bases extracted with out method is improved compared to the baseline.}
    \label{fig:resnet_ablation_t0}
\end{figure}

\begin{figure}
    \centering
    \includegraphics[width=3.4in]{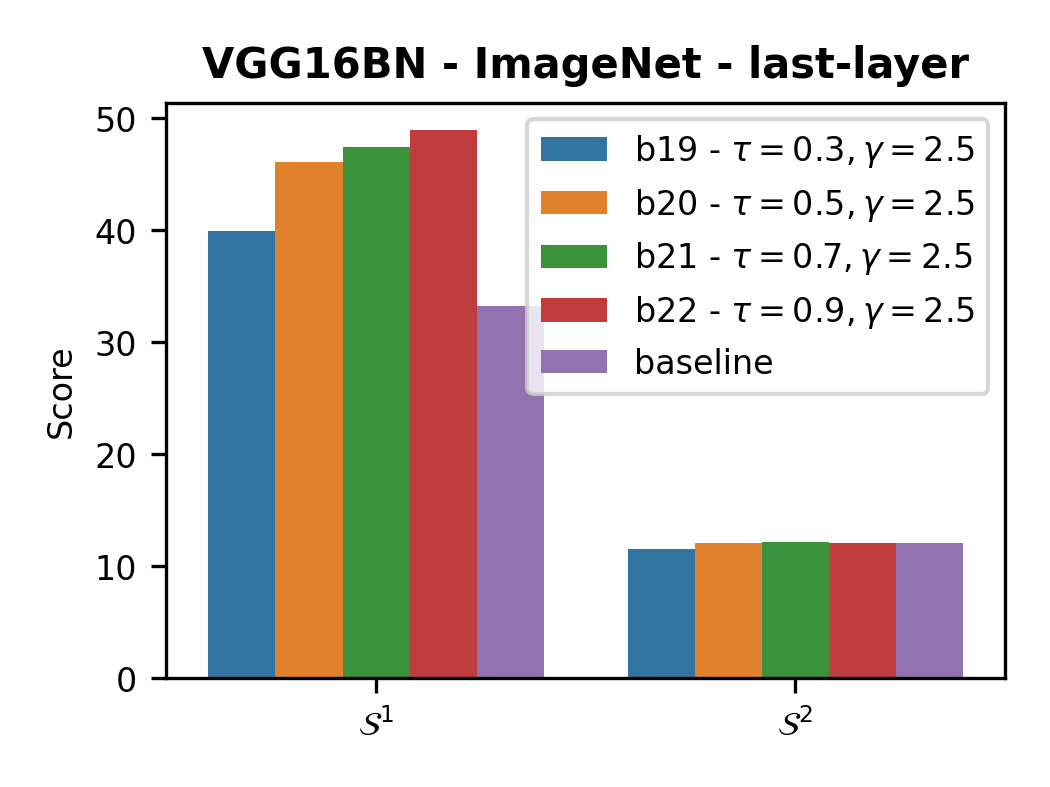}
    \caption{Ablation with respect to $\tau$. With the addition of ICL the interpretability of the bases extracted with out method is improved compared to the baseline, at least for $\Score1$. For $\Score2$, the interpretability of the same bases are comparable to the interpretability of the baseline.}
    \label{fig:vgg16bn_ablation_t0}
\end{figure}

\textbf{Ablating $\tau$ with ICL}
In the second ablation, we make use of \textit{ICL} and set $\lambda^{mm} = 0.5$ and $\lambda^{ic}=5$. For comparisons, we vary $\tau$ to take values from the set $\{0.3, 0.5, 0.7, 0.9\}$. In this study we also use one partition ($N=1, \alpha_{0} = 1, \omega_{0} = 1$) and set $\gamma = 2.5$. Basis interpretability scores are given in Fig. \ref{fig:resnet_ablation_t0} and \ref{fig:vgg16bn_ablation_t0}. The first observation for $\Score1$ is that, in contrast to the previous ablation and for both networks, the extracted bases are significantly more interpretable compared to the baseline. This, experimentally demonstrates the importance of ICL to obtain a meaninful basis. For $\Score2$, a notable improvement over the baseline is provided for ResNet18, while for VGG16BN the interpretability of all the bases, regardless of $\tau$, are comparable to the baseline. Overall, regarding $\Score1$, the value of $\tau$ seemed to have larger impact on the bases learned for VGG16BN compared to the bases that were learned for ResNet18, with increasing values of $\tau$ resulting into larger interpretability scores. We think it is reasonable to believe, that this behaviour possibly indicates that the impact of $\tau$ on the basis interpretability results also depends on the network architecture, the dataset used to train it and its relation with the concept dataset that was used to learn the basis. 


\begin{figure}
    \centering
    \includegraphics[width=3.4in]{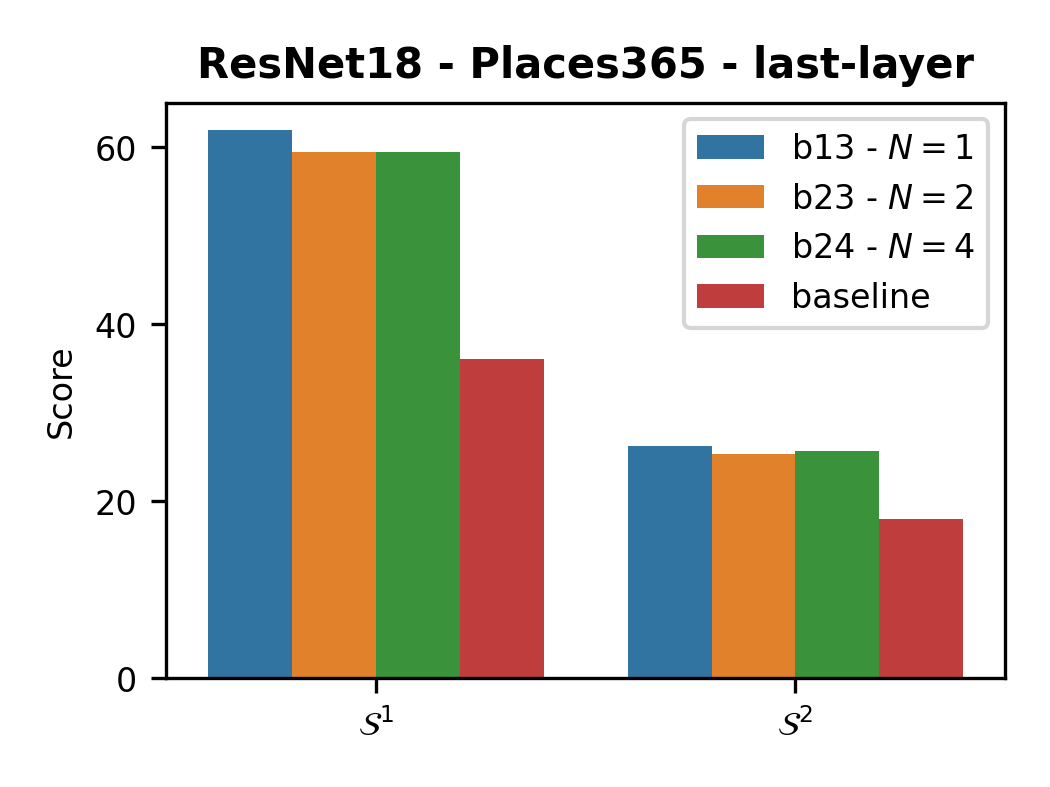}
    \caption{Ablation with respect to the number of partitions $N$. For ResNet18, just a single partition resulted into the most interpretable basis. However, for other values of $N$ the results are comparable and an improvement is noted compared to the baseline.}
    \label{fig:resnet_ablation_partitions}
\end{figure}

\begin{figure}
    \centering
    \includegraphics[width=3.4in]{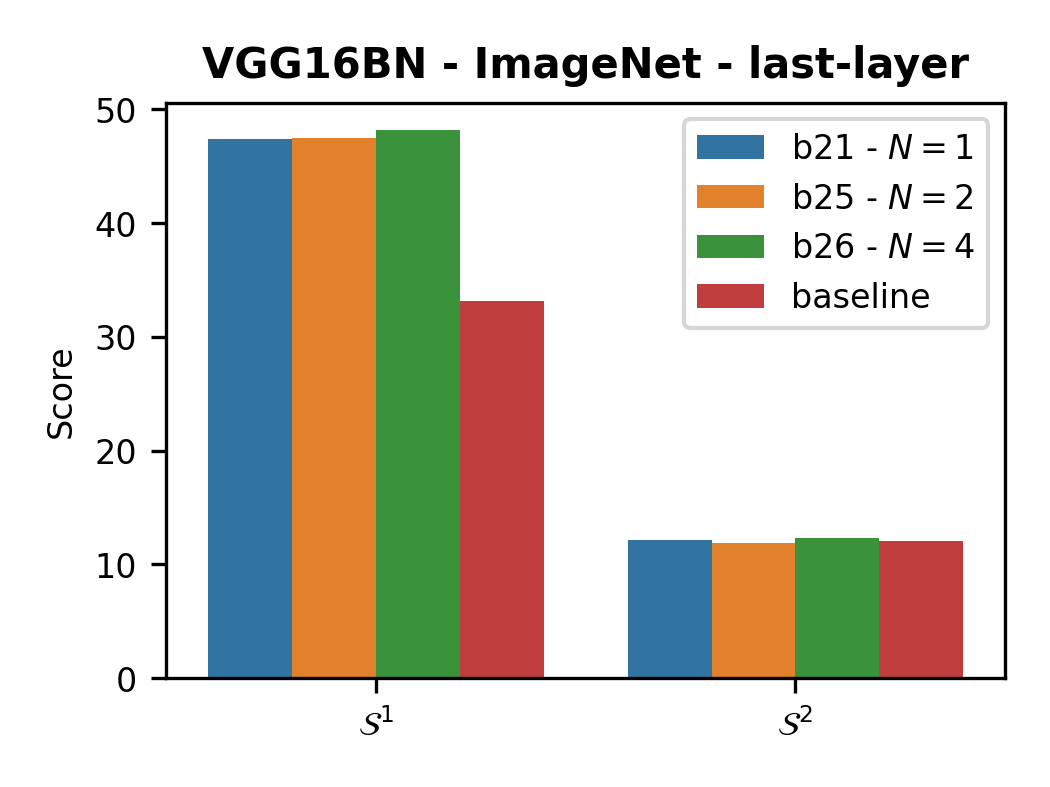}
    \caption{Ablation with respect to the number of partitions $N$. For VGG16BN, four partitions resulted into the most interpretable basis. However, for other values of $N$ the results are comparable and an improvement is noted compared to the baseline.}
    \label{fig:vgg16bn_ablation_partitions}
\end{figure}

\textbf{Ablating partition count with ICL}
In the last ablation, we study the effect of partition count to the basis interpretability scores. In these experiments we considered two cases with different number of partitions (Section \ref{sec:proposed_method}). In both cases, given the number of partitions $N$, we used the following hyper-parameters: $\alpha_{\mu} = 1/N, \mu = \{0,1,...,N-1\}$, and $\omega_{\mu} = \mu+1$. In particular, for the first case we used two partitions ($N=2$) with $\alpha_{\mu} = 0.5, \, \mu\in \{0,1\}$, $\omega_0 = 1, \, \omega_1 = 2$ and in the second case we used four partitions ($N=4$ with $\alpha_{\mu} = 0.25, \mu \in \{0,1,2,3\}$ and $\omega_0 = 1, \, \omega_1 = 2, \, \omega_2 = 3, \, \omega_3 = 4$. In these experiments we used $\tau = 0.7$ and $\gamma = 2.5$. Interpretability results are provided in Fig. \ref{fig:resnet_ablation_partitions}, \ref{fig:vgg16bn_ablation_partitions}. Regarding ResNet18 (Fig. \ref{fig:resnet_ablation_partitions}) we observe that using a single partition ($N=1$) slightly improves the interpretability metrics among the bases that were learned with a larger number of partitions. For VGG16BN, the same slight improvement applies for the basis that was learned with $N=4$. Overall, we could say, that on those experiments and for the given interpretability metrics, the sensitivity of the method with respect to partition count is rather low.

\subsection{Results for more networks}
\begin{figure}
    \centering
    \includegraphics[width=3.4in]{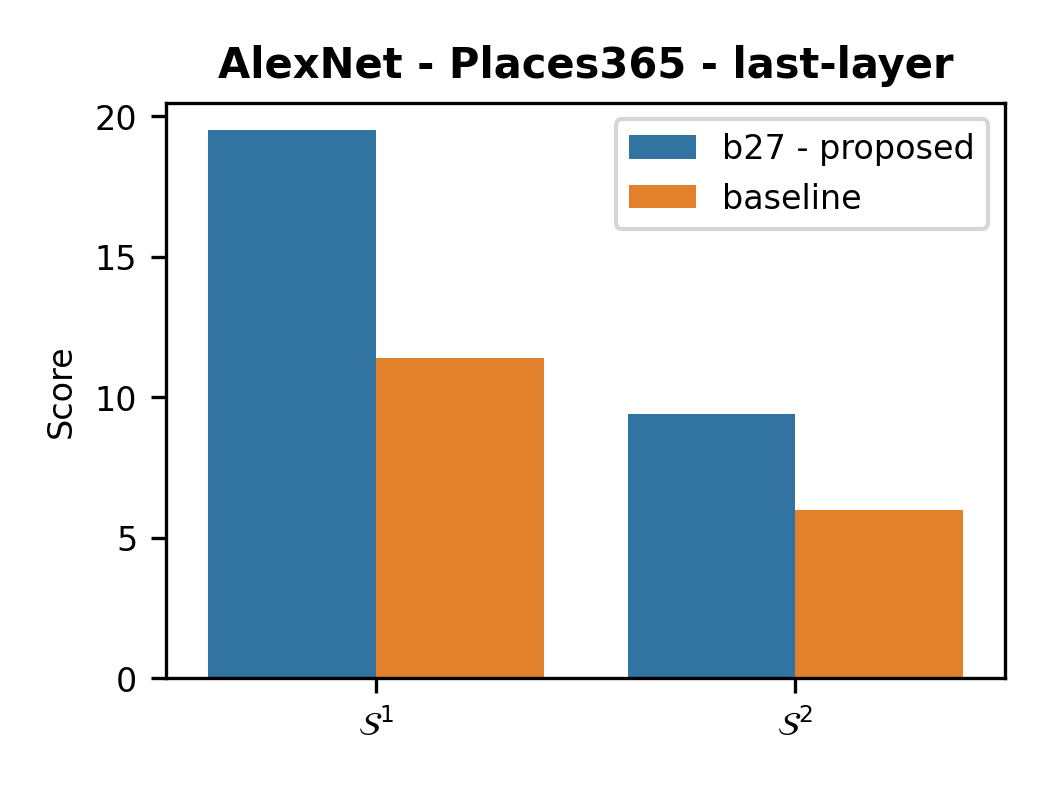}
    \caption{Interpretability comparison between the baseline and a basis extracted with the proposed method. The proposed method suggested a more interpretable basis than the baseline.}
    \label{fig:alexnet}
\end{figure}

\begin{figure}
    \centering
    \includegraphics[width=3.4in]{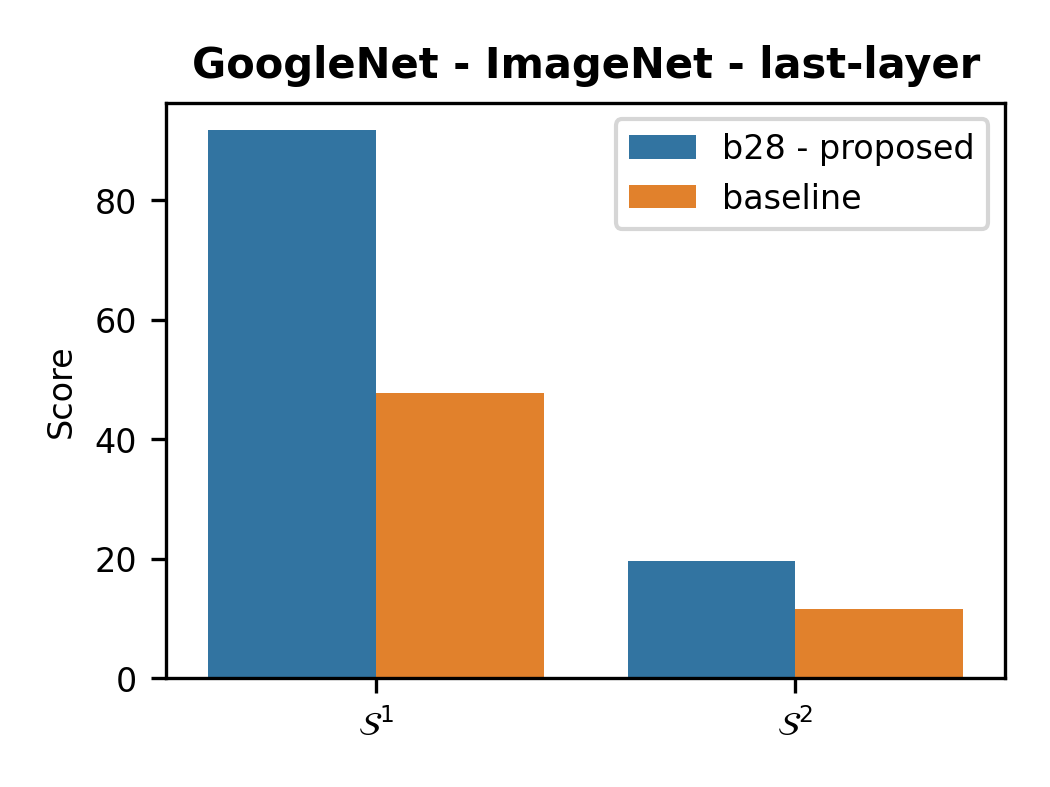}
    \caption{Interpretability comparison between the baseline and a basis extracted with the proposed method. The proposed method suggested a more interpretable basis than the baseline.}
    \label{fig:googlenet}
\end{figure}

In this section we apply the proposed method for interpretable basis extraction to two more networks. We consider AlexNet \cite{NIPS2012_Krizhevsky_AlexNet} (trained on Places365) and GoogleNet \cite{Szegedy_GoogleNet_2014} (trained on ImageNet). Regarding hyper-parameters, we use the loss weight factors of Table \ref{tab:weights}, $\tau = 0.7$, $N=1, \, \alpha_{0} = 1.0, \omega_{0} = 1.0, \, \gamma = 2.5$. We provide basis interpretability results that show improvement over the baseline in Fig. \ref{fig:alexnet} and \ref{fig:googlenet}.


\subsection{Comparison with a supervised approach}
\label{sec:experimental_supervised}
\begin{figure}
    \centering
    \includegraphics[width=3.4in]{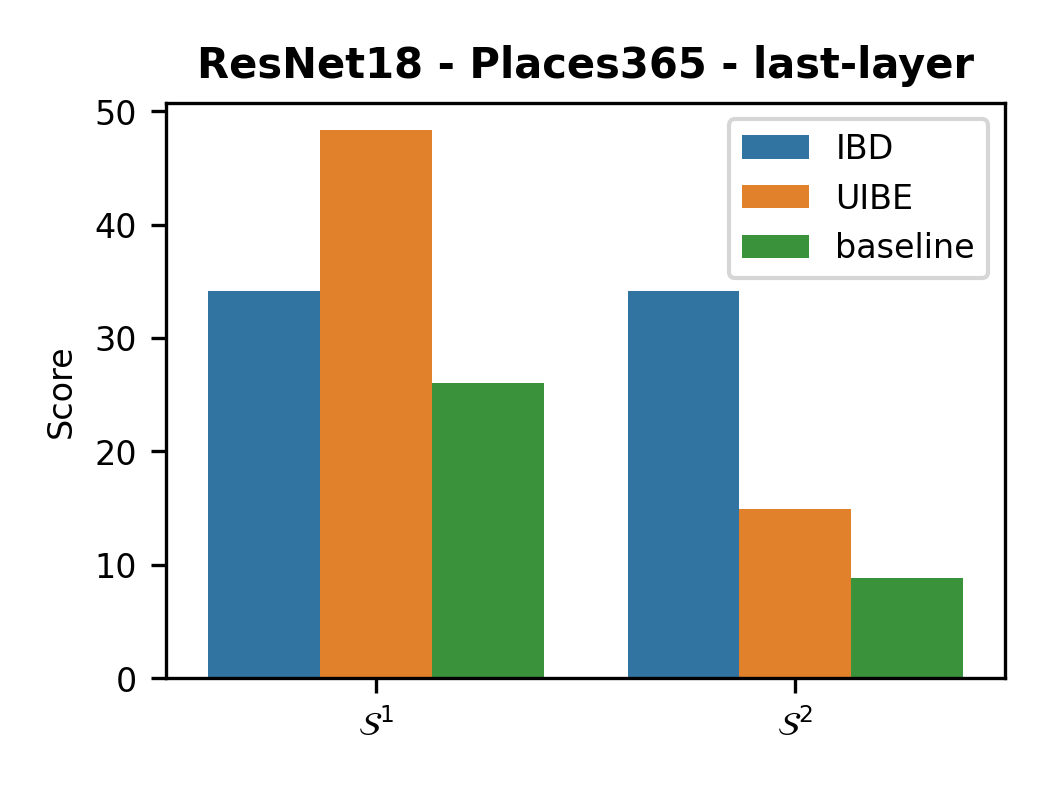}
    \caption{Comparing basis interpretability of the proposed method (\textit{UIBE}) with the natural feature space basis (\textit{baseline}) } and a basis extracted with a supervised approach (\textit{IBD}) \cite{Zhou_Interpretable_Basis_Decomposition}.
    \label{fig:resnet18_ibd}
\end{figure}

\begin{figure}
    \centering
    \includegraphics[width=3.4in]{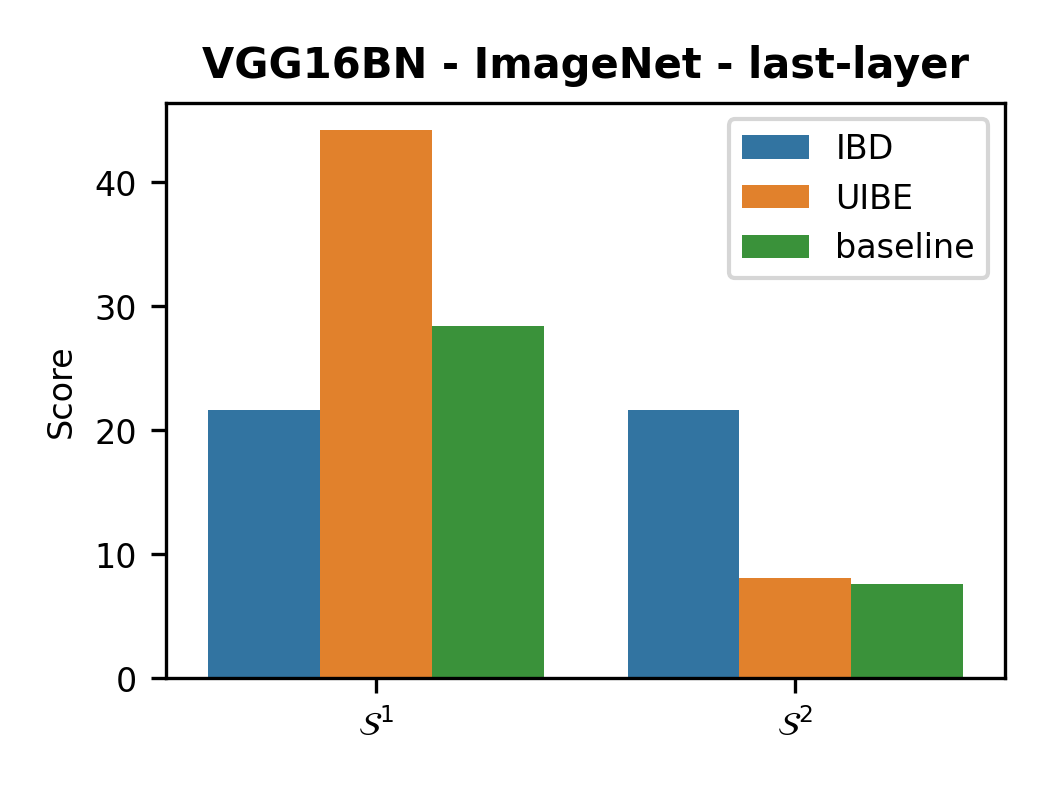}
    \caption{Comparing basis interpretability of the proposed method (\textit{UIBE}) with the natural feature space basis (\textit{baseline}) } and a basis extracted with a supervised approach (\textit{IBD}) \cite{Zhou_Interpretable_Basis_Decomposition}.
    \label{fig:vgg16bn_ibd}
\end{figure}

In this section we compare the interpretability of bases extracted with the proposed method against the \textit{baseline} and the supervised approach of \cite{Zhou_Interpretable_Basis_Decomposition}. Once again, we consider the \textit{last-layers} of ResNet18 (trained on Places365) and VGG16BN (trained on ImageNet). We followed the approach of Interpretable Basis Decomposition (IBD) \cite{Zhou_Interpretable_Basis_Decomposition} and learned a basis in a supervised way for the concepts categories of \textit{objects} and \textit{parts}. To learn the basis we used the training split of the concept dataset. The number of basis vectors that were learned from IBD was $I=660$ while the dimensionality of the feature space for both CNNs is $D=512$. Regarding the proposed method, for ResNet18, we re-considered the basis \textit{b13} (Fig. \ref{fig:resnet_ablation_t0}) which was learned from all images (regardless the category annotations) of the concept dataset's training split. This time though, we only considered the concept categories of \textit{objects} and \textit{parts} to label the basis. We did the same for the natural feature space basis as well. Finally we report $\Score1$ and $\Score2$ on the validation split of the same dataset. A similar approach was taken for VGG16BN, where we re-considered the basis \textit{b26}. The results for the two networks are given in Fig. \ref{fig:resnet18_ibd} and \ref{fig:vgg16bn_ibd}. 

From the previously mentioned figures, we first observe, that the bases learned with IBD have the same score on both metrics. This is actually expected, since all concept labels in a basis learned with IBD are unique. On the contrary, when labeling the natural feature space basis or a basis extracted with the proposed method, the same concept label may be attributed to more than one basis vectors. This also might be a possible explanation for why the proposed method showcases significantly better interpretability scores for $\Score1$ compared to IBD. In other words, IBD is limited to learn a single direction for each one of the concepts, while the bases extracted with the proposed method may cover more than one direction for the same concept. Additionally, the sparsity criterion which we use to learn the interpretable basis, ensures that the different basis vectors cover different parts of the concept dataset. Another factor to consider for the same matter is the basis labeling procedure, which in our case is \cite{bau_NetDissection_CVPR}. Other basis labeling strategies might suggest different labels which might also affect the interpretability scores. It is also noteworthy that the same possible explanation might be given regarding Fig. \ref{fig:vgg16bn_ibd} where even the natural feature space basis scores better than IBD in terms of $\Score1$. 

Regarding $\Score2$, the bases extracted with IBD may be considered significantly more interpretable than the bases extracted with the current work, with the latter being even more prominent in the case of ResNet18. We think that this fact is also linked to the previous argument. In particular, since a basis learned without supervision may have duplicate labels, the number of unique concept labels that can be attributed to the basis vectors (which is related to what $\Score2$ measures) is expected to be less than the number of vectors in the basis. However, for a basis learned in a supervised way, these two numbers are always equal. Moreover, in this case, IBD used a basis with a larger number of vectors ($I=660$) compared to the proposed method (which only uses $I=D=512$).

Overall, we find it difficult to strictly position the proposed method in relation to a supervised approach for this problem. We think that the current work reveals a possible limitation of the supervised approach which assumes that concept representations lie only on a single direction of the feature space. The proposed method has the potential to overcome this limitation. However, the presented experimental results also suggest that the previously mentioned strength of the proposed method is also its limitation. By devoting more than one basis direction to a single concept, inevitably limits the number of different unique concepts that can described by the basis. A possible direction towards improvement might be to consider an \textit{approximately} orthogonal and over-complete basis of the feature space (i.e. $I>D$), which we leave for future work.

\subsection{Qualitative comparisons}
\begin{figure}
    \centering
    \includegraphics[width=3.4in]{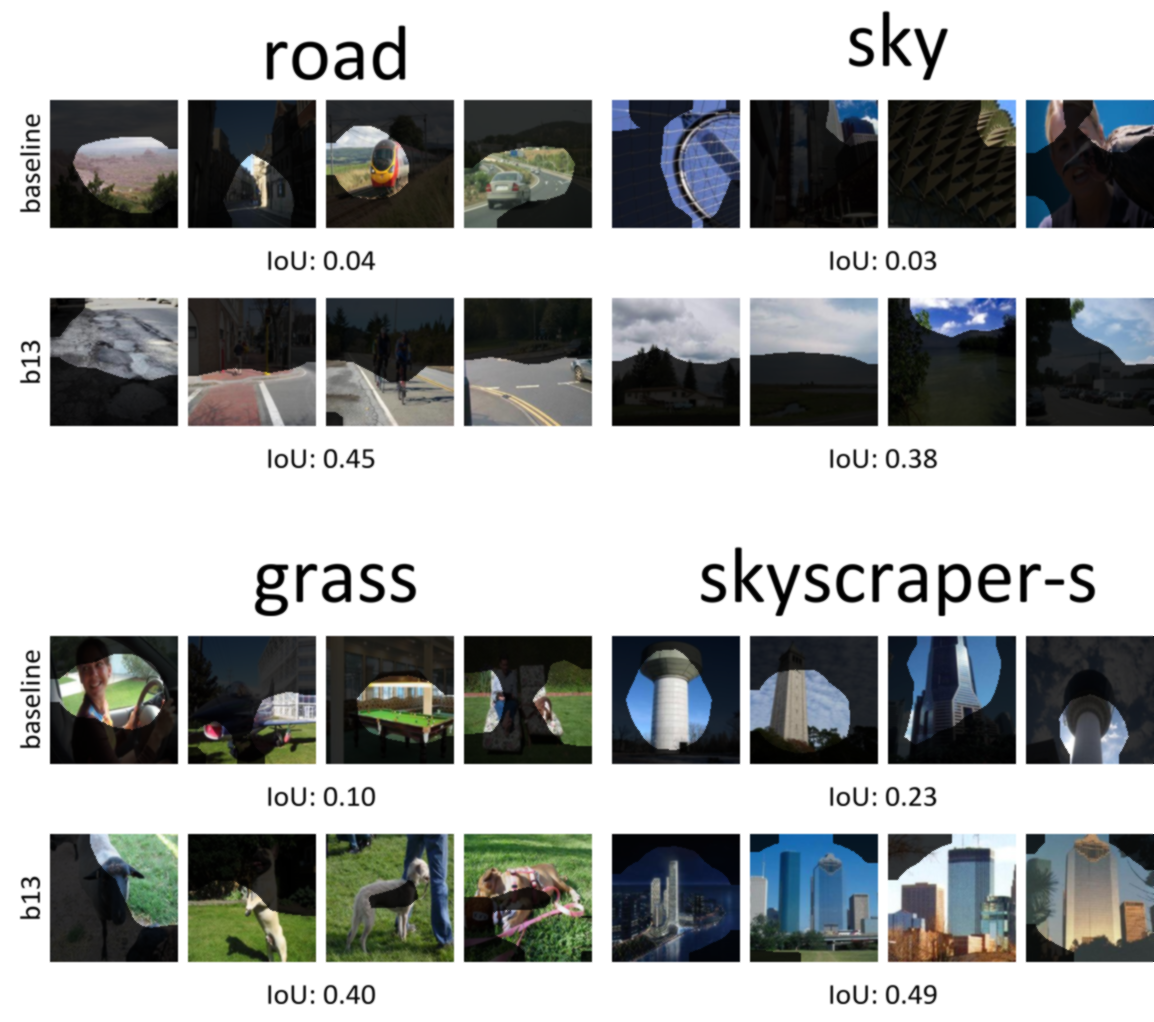}
    \caption{Qualitative results for ResNet18.}
    \label{fig:qualitative_resnet1}
\end{figure}

\begin{figure}
    \centering
    \includegraphics[width=3.4in]{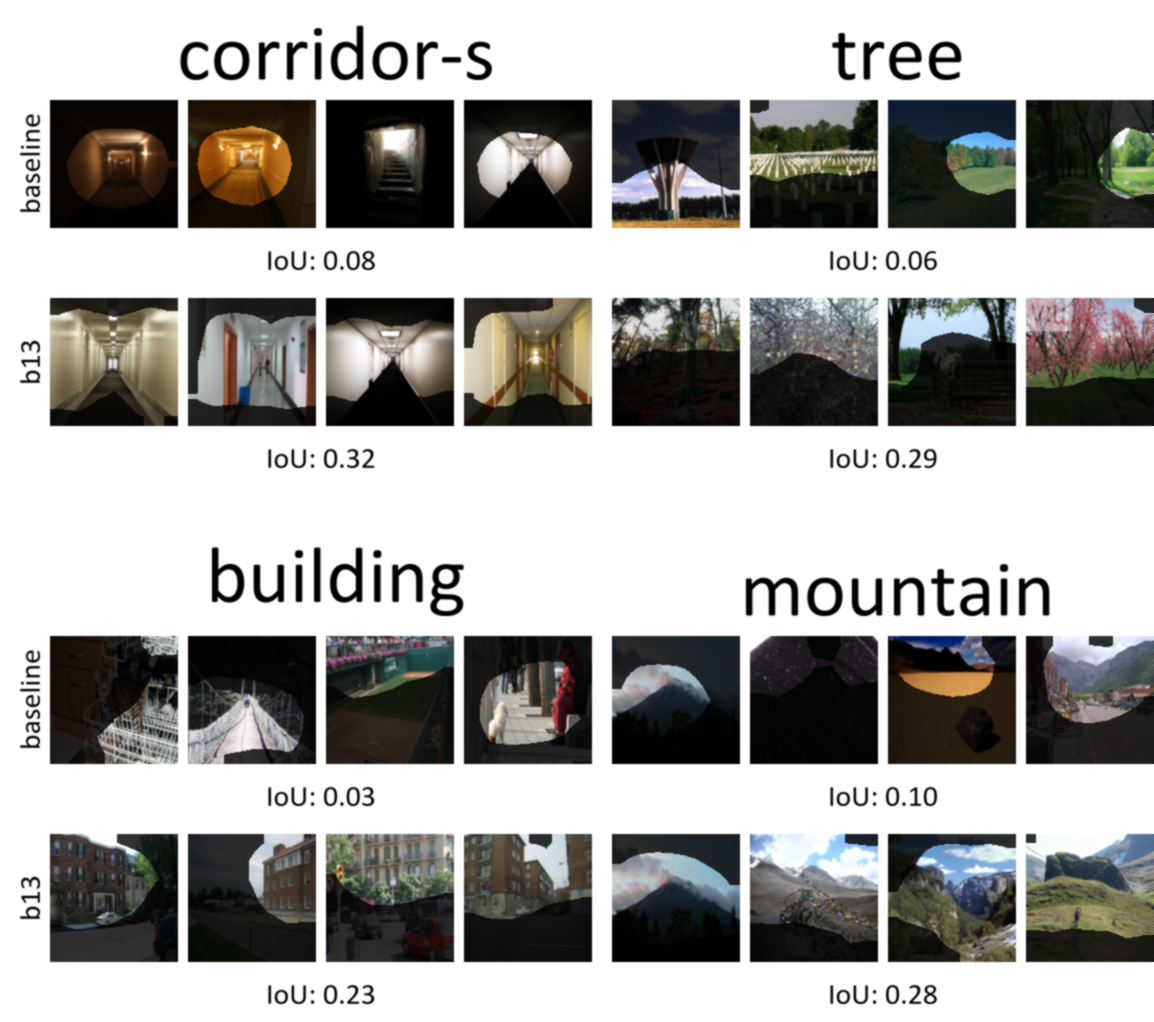}
    \caption{Qualitative results for ResNet18.}
    \label{fig:qualitative_resnet2}
\end{figure}

\begin{figure}
    \centering
    \includegraphics[width=3.4in]{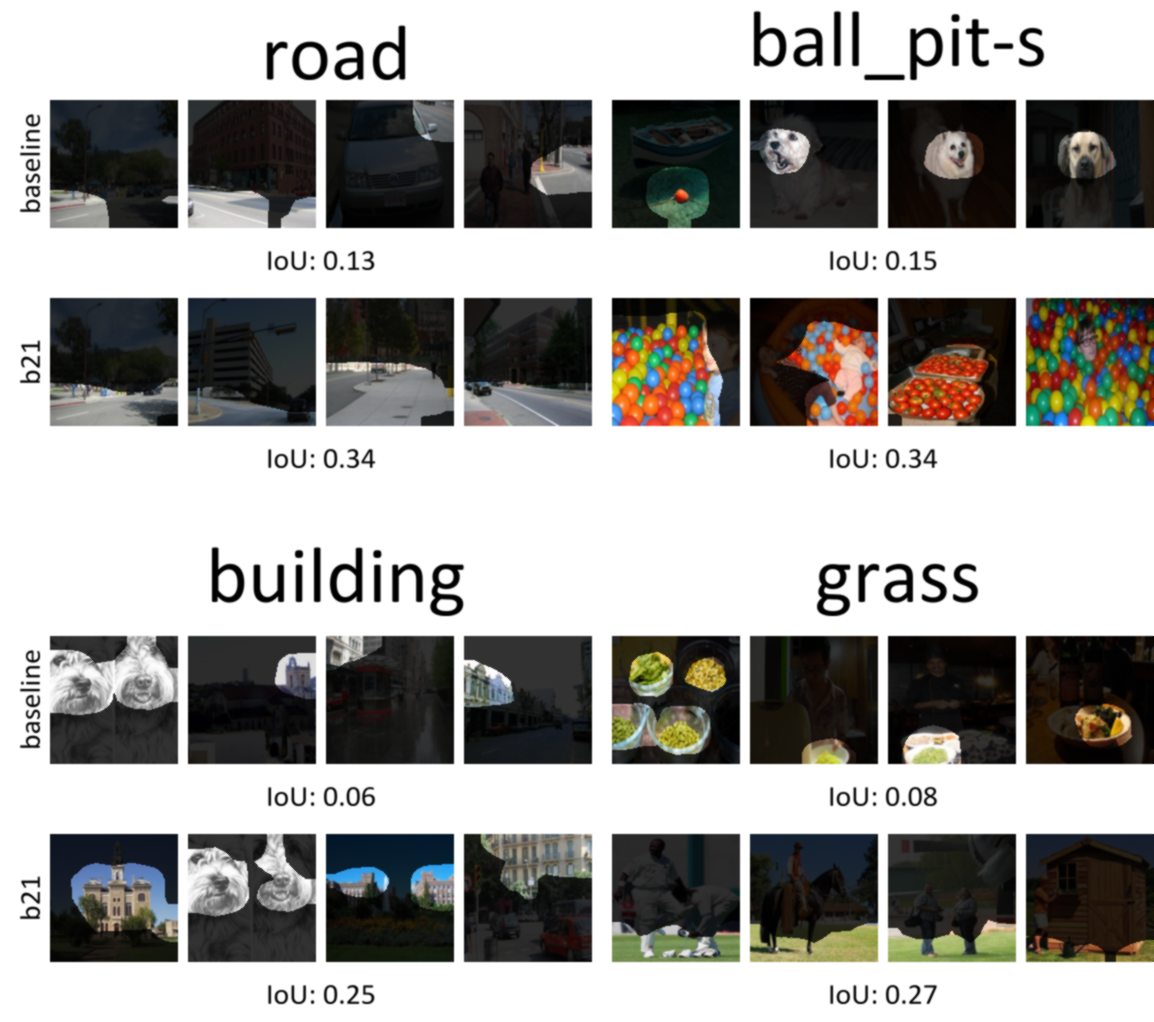}
    \caption{Qualitative results for VGG16BN.}
    \label{fig:qualitative_vgg16bn1}
\end{figure}

\begin{figure}
    \centering
    \includegraphics[width=3.4in]{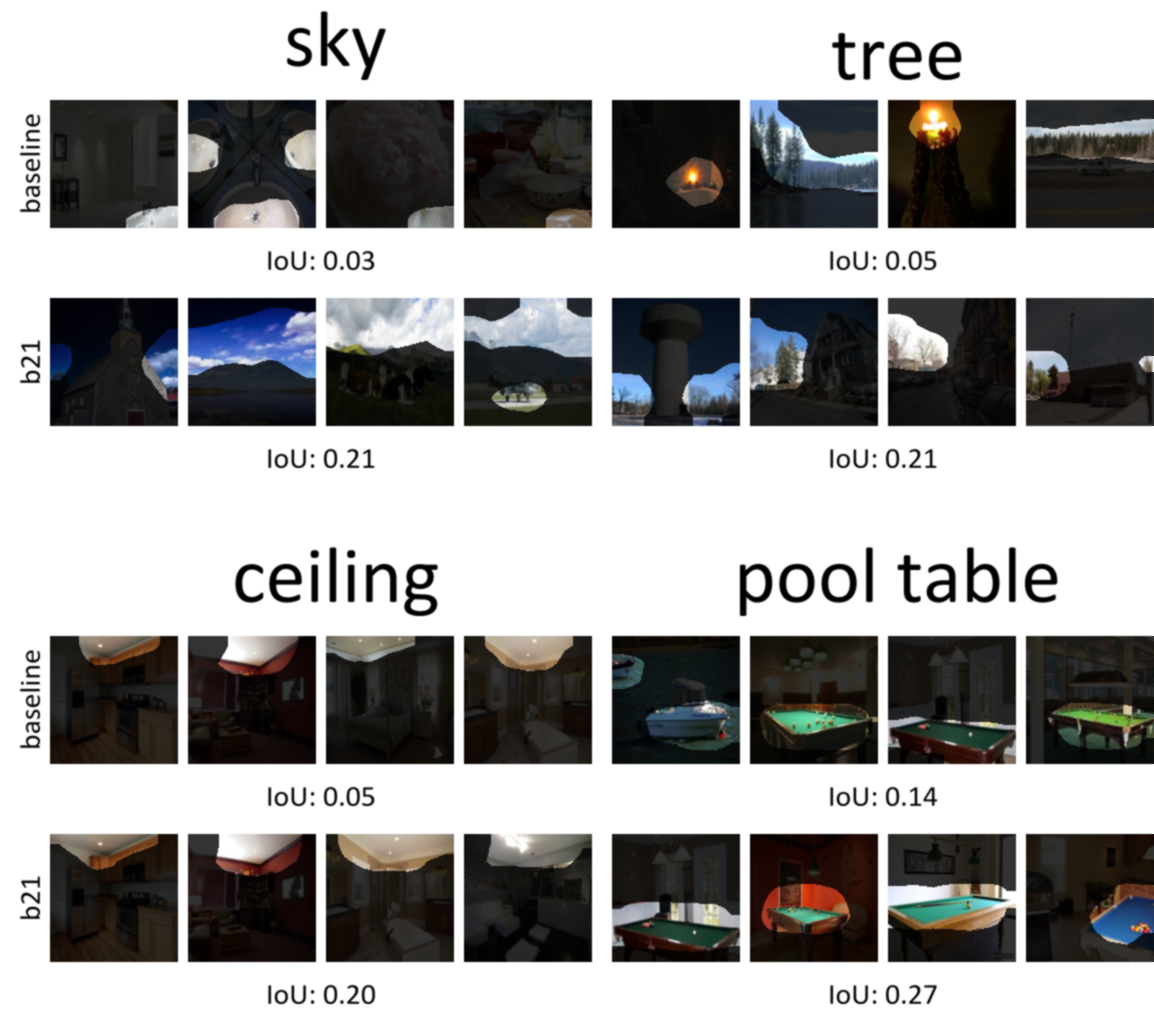}
    \caption{Qualitative results for VGG16BN.}
    \label{fig:qualitative_vgg16bn2}
\end{figure}

\begin{figure}
    \centering
    \includegraphics[width=3.4in]{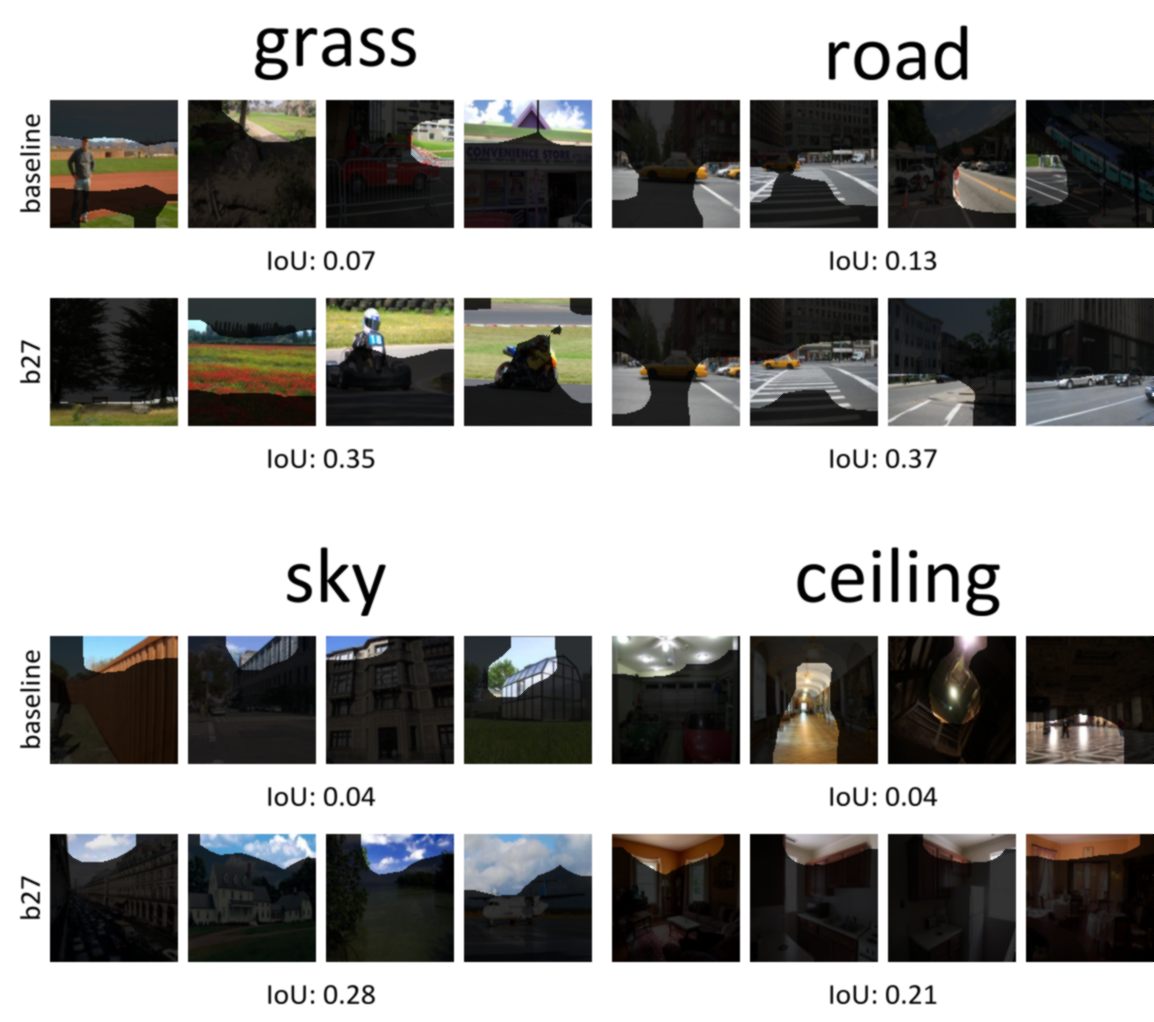}
    \caption{Qualitative results for AlexNet.}
    \label{fig:qualitative_alexnet1}
\end{figure}

\begin{figure}
    \centering
    \includegraphics[width=3.4in]{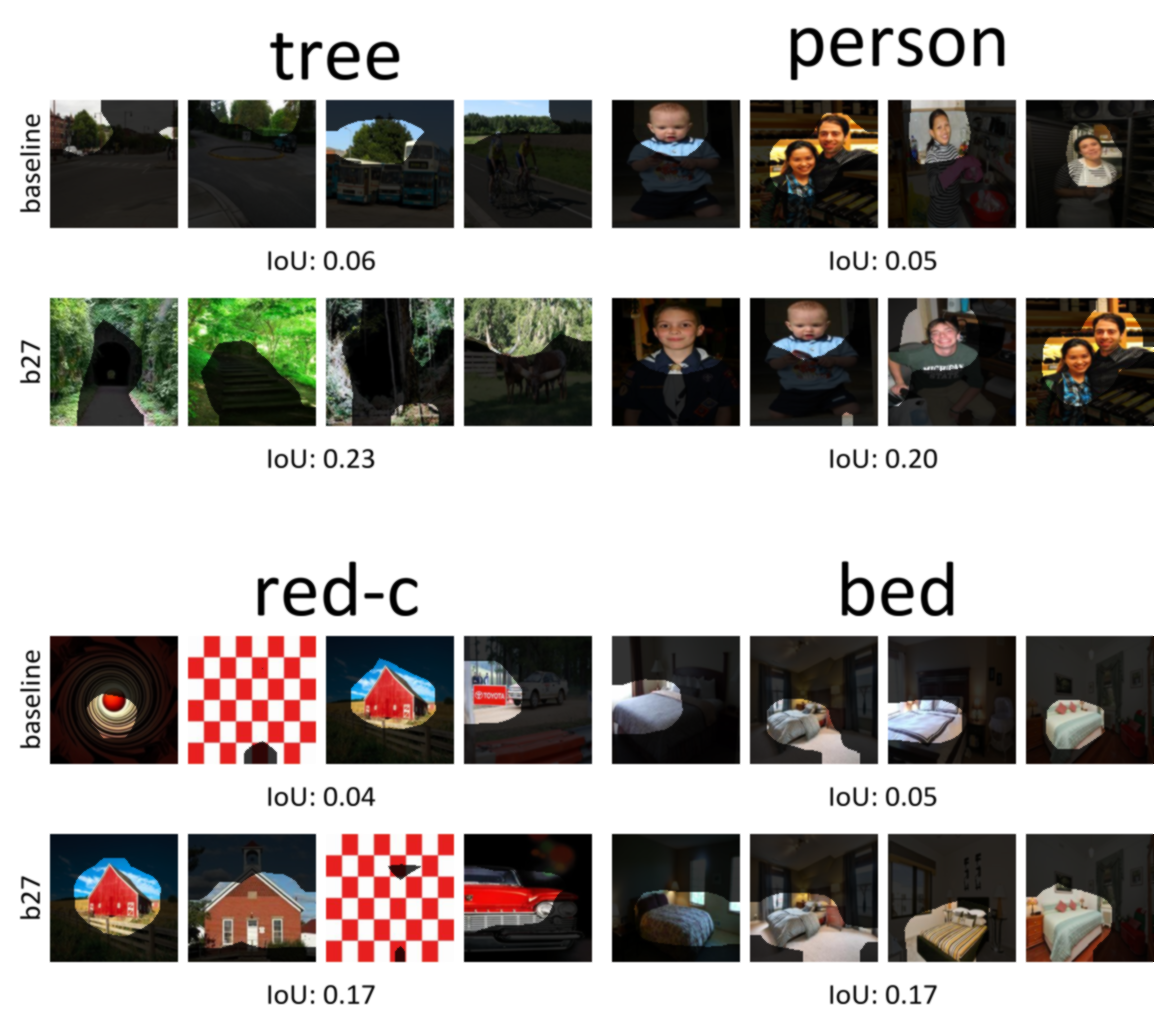}
    \caption{Qualitative results for AlexNet.}
    \label{fig:qualitative_alexnet2}
\end{figure}

\begin{figure}
    \centering
    \includegraphics[width=3.4in]{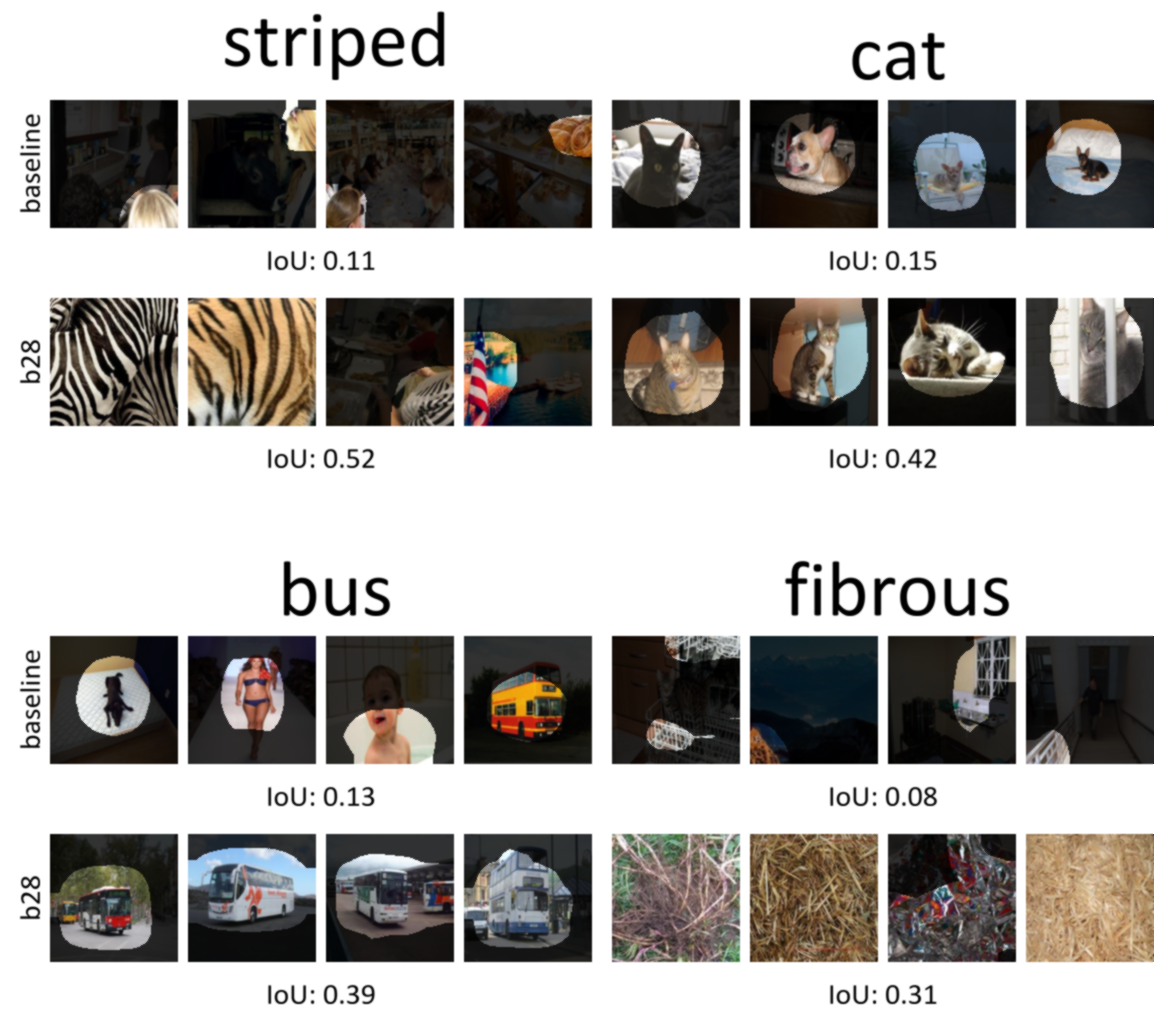}
    \caption{Qualitative results for GoogleNet.}
    \label{fig:qualitative_googlenet1}
\end{figure}

\begin{figure}
    \centering
    \includegraphics[width=3.4in]{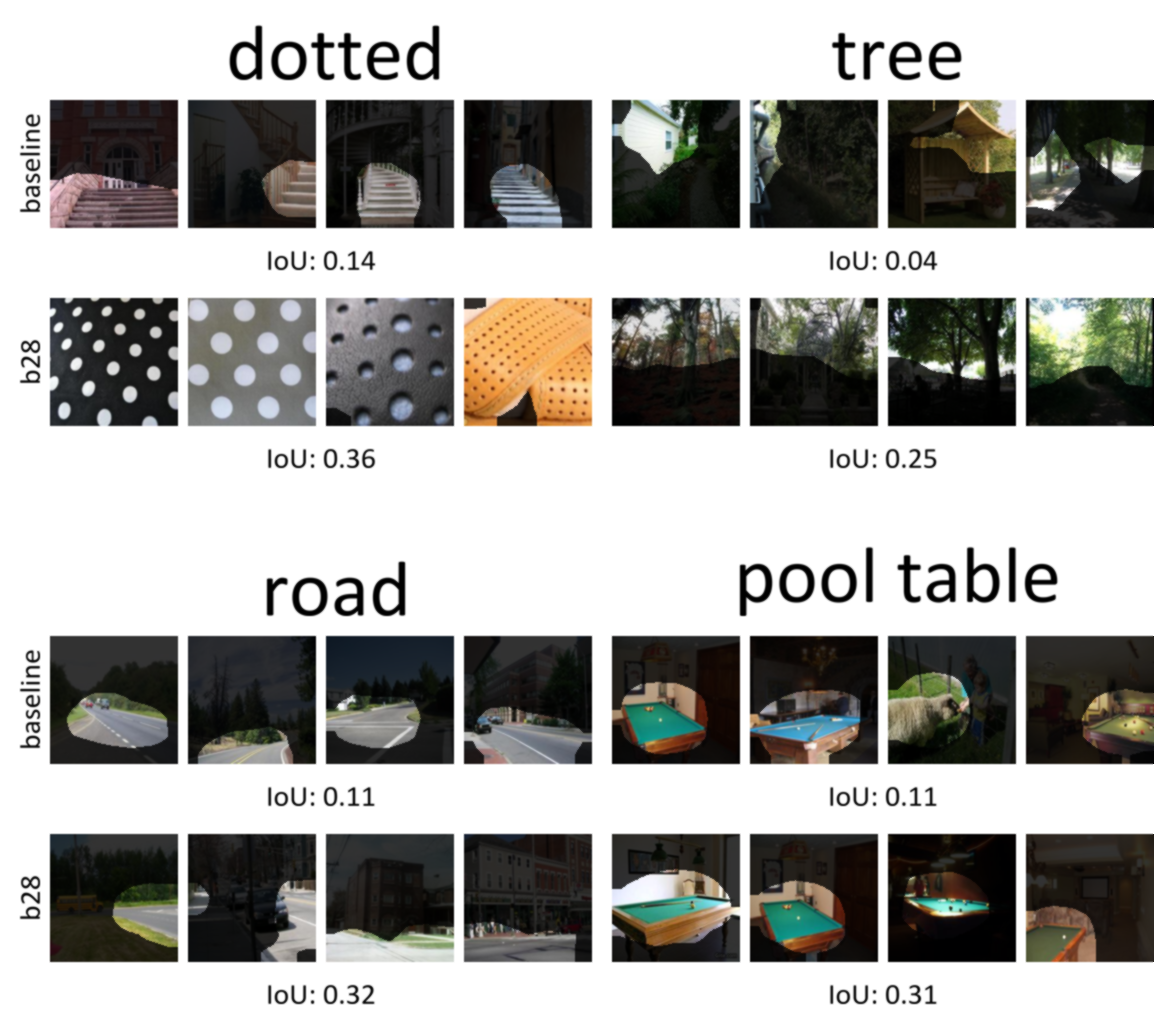}
    \caption{Qualitative results for GoogleNet.}
    \label{fig:qualitative_googlenet2}
\end{figure}

In this section we provide qualitative results which highlight the interpretability improvement gains that are obtained when we transform image feature representations to a basis learned with the proposed method. Thus, Fig. \ref{fig:qualitative_resnet1} and \ref{fig:qualitative_resnet2} depict results for ResNet18, Fig. \ref{fig:qualitative_vgg16bn1} and \ref{fig:qualitative_vgg16bn2} for VGG16BN, Fig. \ref{fig:qualitative_alexnet1} and \ref{fig:qualitative_alexnet2} for AlexNet and Fig. \ref{fig:qualitative_googlenet1} and \ref{fig:qualitative_googlenet2} for GoogleNet. In those figures, we used \cite{bau_NetDissection_CVPR} to assign concept labels for the bases vectors extracted by our method, as well as to the vectors of the natural feature space bases. Among the group of common concepts that have been assigned to the concept detectors of the two bases, we considered the top-performing concept detector in each basis. The common name of the concept detectors is given on the (sub-)figure's top. The basis name that each concept detector comes from, is written on the left. For each selected concept detector, we present a row of images whose representations have a spatial element which is ranked among the top-4 activations over the validation split of the concept dataset ($\Kvl$). The reported IoU scores which are given below the set of images, corresponds to the IoU performance of the respective concept detector over the whole set of images in the validation set of the concept dataset. Each figure is meant to be read as a $2 \times 2$ grid of 4 concepts with each cell containing $2 \times 4$ images.

\subsection{Are the bases learned with a supervised approach orthogonal ?}

In this last section of experimental results, we experimentally seek to validate our hypothesis that an interpretable basis should be orthogonal. While our hypothesis is based on the assumption that the CNN has linearly disentangled concept representations, we still try to, at least partially, answer to what extend this is already happening when we use a supervised method to learn an interpretable basis. Building on our previous experimental results, we consider the bases that we learned with IBD \cite{Zhou_Interpretable_Basis_Decomposition} for the \textit{last-layers} of ResNet18 and VGG16BN. We provide statistical measurements for the distribution of angles between basis vectors that are met in those bases. The results are depicted in Table \ref{tab:ibd_basis_angles}. It is noteworthy to mention that those bases have $I=660$ and $D=512$, with the important relation that $I>D$. Based on our measurements, the bases could be considered approximately orthogonal since the mean angle between any pairs of basis vectors is around $86.5^\circ$ and the standard deviation of the distribution is less than $3.77^\circ$, in the worst case. This fact could further support our intuition that interpretable bases shall be orthogonal. While the present work considers only $I\leq D$, future extensions could study the case where $I \ge D$ with approximate orthogonality constraints.
 
\begin{table}
\centering
\caption{
Statistics of Pairwise Vector Angles for the bases that we learned with the supervised approach of IBD \cite{Zhou_Interpretable_Basis_Decomposition}.
}
\label{tab:ibd_basis_angles}
\resizebox{1.0\columnwidth}{!}{
\begin{tabular}{ccccc}
\cline{2-5}
\multicolumn{1}{c|}{}                  & \multicolumn{4}{c|}{\textbf{Pairwise Vector Angles (deg)}}                                                                                     \\ \hline
\multicolumn{1}{|c|}{\textbf{Network}} & \multicolumn{1}{c|}{\textbf{Mean}} & \multicolumn{1}{c|}{\textbf{Std}} & \multicolumn{1}{c|}{\textbf{Min}} & \multicolumn{1}{c|}{\textbf{Max}} \\ \hline
\multicolumn{1}{|c|}{ResNet18}         & \multicolumn{1}{c|}{85.67}         & \multicolumn{1}{c|}{3.77}         & \multicolumn{1}{c|}{45.5}         & \multicolumn{1}{c|}{99.5}         \\ \hline
\multicolumn{1}{|c|}{VGG16BN}          & \multicolumn{1}{c|}{88.23}         & \multicolumn{1}{c|}{2.84}         & \multicolumn{1}{c|}{57.71}        & \multicolumn{1}{c|}{100.0}        \\ \hline
\multicolumn{1}{l}{}                   & \multicolumn{1}{l}{}               & \multicolumn{1}{l}{}              & \multicolumn{1}{l}{}              & \multicolumn{1}{l}{}             
\end{tabular}
}
\end{table}
\vspace{-7pt}
\section{Conclusion}
\label{sec:conclsion}
We presented an unsupervised, post-hoc method to extract an interpretable basis for the feature space of a CNN's intermediate layer. Based on current literature, we also proposed two metrics that can be used to measure a basis for its interpretability. We evaluated the effectiveness of the proposed method in standard CNN architectures and demonstrated that intermediate layer representations become more interpretable when projected onto bases extracted with our method. Finally, using the proposed metrics, we compared the outcomes of our method with the outcomes of a method that derives an interpretable basis using supervision. According to the  interpretability metrics, the bases extracted with the proposed method, in one aspect, show appreciable interpretability improvements over the bases extracted with the supervised approach. At the same time, in a second aspect, the bases derived with supervision were significantly more interpretable than the bases that were suggested by our method. This fact might seem peculiar at first. However, a possible explanation was provided and directions for future research for deeper understanding were suggested. We hope that the present work has contributed additional knowledge to interpretable basis extraction and motivates further research for understanding \textit{black-box} models.





\bibliographystyle{IEEEtran}
\bibliography{IEEEabrv,refs.bib}

\begin{IEEEbiography}[{\includegraphics[width=1in,height=1.25in,clip,keepaspectratio]{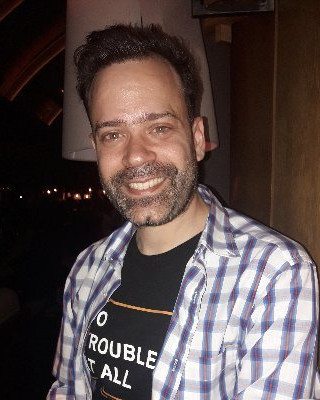}}]{Alexandros Doumanoglou}{\space} received the Diploma in
electrical and computer engineering from Aristotle University of Thessaloniki, Thessaloniki, Greece, in 2009 and joined the Information Technologies Institute, in 2012. Currently, he is working toward
the Ph.D. degree in explainable artificial intelligence at the Department of Advanced Computing Sciences of Maastricht University, The Netherlands.
His current research focuses on unsupervised learning and explainable and interpretable methods for deep learning models.
\end{IEEEbiography}

\begin{IEEEbiography}
[{\includegraphics[width=1in,height=1.25in,clip,keepaspectratio]{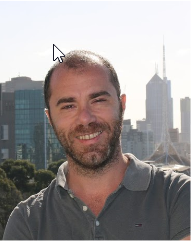}}]{Stylianos Asteriadis}{\space} received the diploma of Electrical and Computer Engineer from Aristotle University of Thessaloniki, Thessaloniki, Greece in 2004, the M.Sc. degree in digital media from the School of Informatics at the same university in
2006, and the Ph.D. in Electrical and Computer Engineering from the National Technical University of Athens, Athens, Greece, in 2011.
He was an Associate Professor at the Department of Advanced Computing Sciences at Maastricht University, Maastricht, The Netherlands, until the
final acceptance of this paper, where he coordinated the Cognitive Systems
Group. He is currently working at the European Commission.\footnote{The information and views set out in this article are those of the authors and do not necessarily reflect the official opinion of the Institution.} 
\end{IEEEbiography}

\begin{IEEEbiography}[{\includegraphics[width=1in,height=1.25in,clip,keepaspectratio]{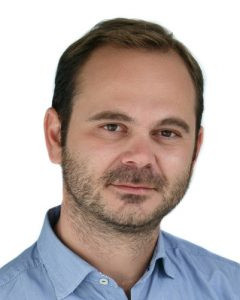}}]{Dimitrios Zarpalas}{\space} received the Diploma in electrical and computer engineering from Aristotle University of Thessaloniki (A.U.Th), Thessaloniki, Greece in 2003, the M.Sc. degree in electrical engineering from the Pennsylvania State University, Philadelphia, USA, in 2006, and the Ph.D. degree in medical informatics from the Department of Medicine, Health Science School, A.U.Th, in 2014.
He joined the Information Technologies Institute, Thessaloniki, Greece, in 2007, where he is currently a Researcher, grade B. His research interests include tele-immersion applications, 3-D computer vision, 3-D object recognition, and motion capturing.
\end{IEEEbiography}

\end{document}